\documentclass[10pt,twocolumn,letterpaper]{article}

\usepackage{wacv}
\usepackage{times}
\usepackage{epsfig}
\usepackage{graphicx}
\usepackage{amsmath}
\usepackage{amssymb}

\usepackage[flushleft]{threeparttable}
\usepackage{balance}

\wacvfinalcopy 

\ifwacvfinal
\def\assignedStartPage{1} 
\fi

\ifwacvfinal
\usepackage[breaklinks=true,bookmarks=false]{hyperref}
\else
\usepackage[pagebackref=true,breaklinks=true,colorlinks,bookmarks=false]{hyperref}
\fi

\ifwacvfinal
\else
\fi

\begin{document}

\title{MECCANO: A Multimodal Egocentric Dataset for Humans Behavior Understanding in the Industrial-like Domain}

\author{Francesco Ragusa, Antonino Furnari, Giovanni Maria Farinella\\
FPV@IPLAB - University of Catania\\
Next Vision s.r.l., Spin-off of the University of Catania\\
}

\maketitle

\begin{abstract}
Wearable cameras allow to acquire images and videos from the user's perspective. These data can be processed to understand humans behavior. 
Despite human behavior analysis has been thoroughly investigated in third person vision, it is still understudied in egocentric settings and in particular in industrial scenarios.
To encourage research in this field, we present MECCANO, a multimodal dataset of egocentric videos to study humans behavior understanding in industrial-like settings. The multimodality is characterized by the presence of gaze signals, depth maps and RGB videos acquired simultaneously with a custom headset.
The dataset has been explicitly labeled for fundamental tasks in the context of human behavior understanding from a first person view, such as recognizing and anticipating human-object interactions.
With the MECCANO dataset, we explored five different tasks including 1) Action Recognition, 2) Active Objects Detection and Recognition, 3) Egocentric Human-Objects Interaction Detection, 4) Action Anticipation and 5)~Next-Active Objects Detection. We propose a benchmark aimed to study human behavior in the considered industrial-like scenario which demonstrates that the investigated tasks and the considered scenario are challenging for state-of-the-art algorithms.
To support research in this field, we publicy release the dataset at \url{https://iplab.dmi.unict.it/MECCANO/}.
\end{abstract}

\section{Introduction}
\label{sec:intro}
\begin{figure}[t]
\centering
\includegraphics[width=\columnwidth]{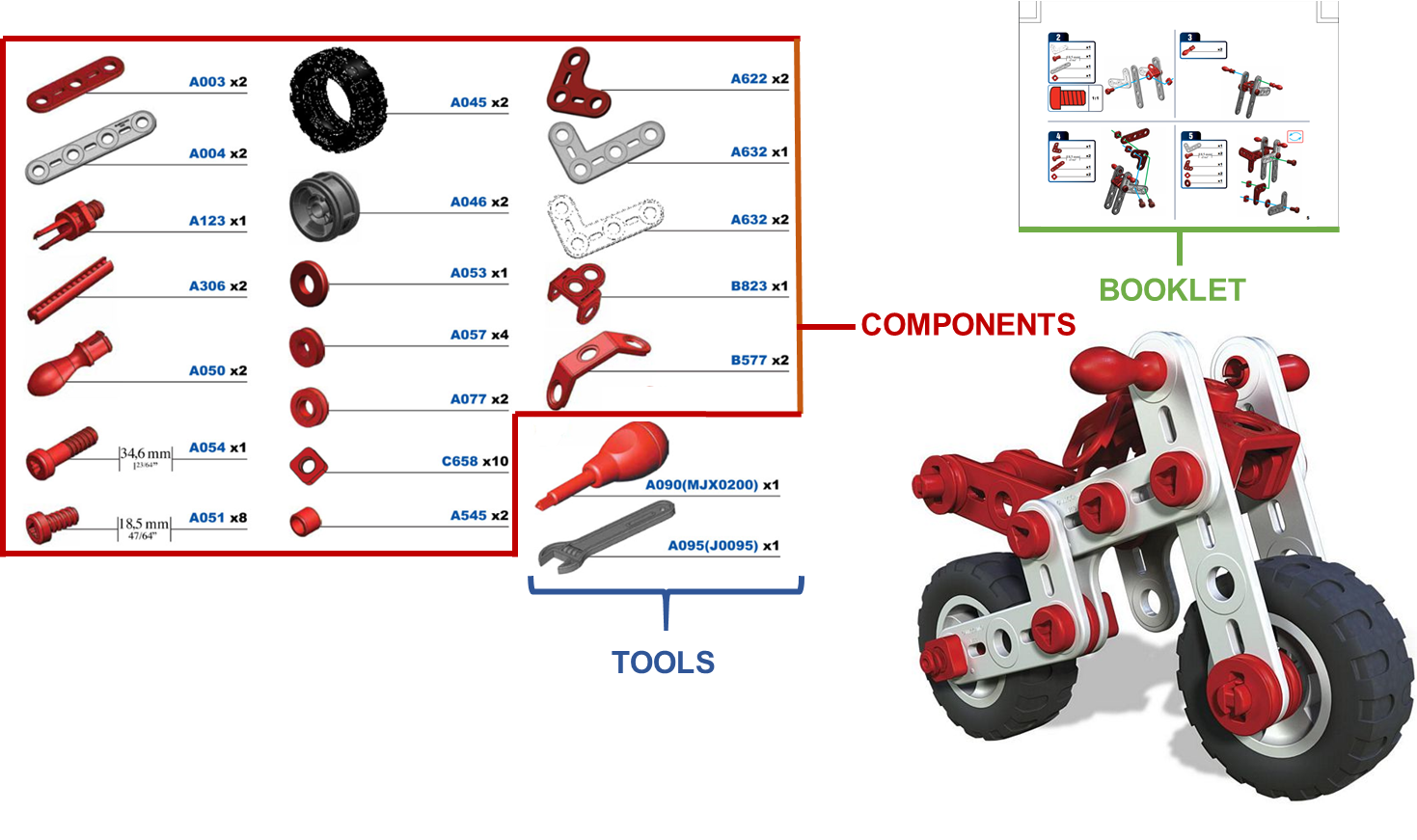}
\caption{Toy model built by subjects interacting with 2 tools, 49 components and the instructions booklet.}
\label{fig:toy_model}
\end{figure}

Understanding human behavior from an egocentric perspective allows to build intelligent systems able to assist humans equipped with a camera (e.g., Microsoft Hololens 2\footnote{\url{https://www.microsoft.com/en-us/hololens}}, Vuzix Blade\footnote{\url{https://www.vuzix.com/products/blade-smart-glasses}}, Nreal Light \footnote{\url{https://www.nreal.ai/light/}}, etc.) in many contexts, including cultural sites \cite{RagusaPRL, cucchiara2014visions, vedi2019}, home scenarios \cite{You-Do_Damen_14} and industrial environments \cite{DeepVisionShield_Colombo19}.

For example, recognizing human-object interactions in an industrial environment from First Person Vision (FPV) can be useful for monitoring the use of machines, to schedule calibration operations, to suggest to the operator how to use a specific machine or object, as well as to issue notifications about actions that may be missed in a production pipeline \cite{miss_actions_shapiro}. Furthermore, anticipating what a worker will do and which objects he will interact with provides information to improve safety in a factory, for example by notifying the user with an alert in case a dangerous action or interaction is anticipated.
Many recent works investigated human behavior understanding considering different tasks such as action recognition \cite{feichtenhofer2018slowfast, TwoStream_convolutional_action_Zisserman_14, Two-Stream_Zisserman, Zhou2018TemporalRR, kazakos2019TBN, TSM_2019}, object detection \cite{girshick2014rich, girshick2015fast, ren2015faster, yolov3}, human-object interaction detection \cite{Gkioxari2018DetectingAR, Gupta2015VisualSR, Hands_in_contact_Shan20, Nagarajan2020EGOTOPOEA}, action anticipation \cite{Felsen_what_will_happen_17, Gao2017REDRE, furnari2019rulstm, slowfast_rulstm_ballan} as well as the detection of the next active objects \cite{Bertasius2017FirstPA, Furnari2017NextactiveobjectPF, JIANG2021212, Ego4D2021}.

Advances in these fields have been obtained thanks to the availability of public datasets \cite{Imagenet, lin2014COCO, Gupta2015VisualSR, HICO_Chao} considering different contexts such as kitchens \cite{Damen2018EPICKITCHENS,Damen2020RESCALING, Li2018_EGTEA-GAZE+, Torre2009CMU-MMAC}, home and offices \cite{Ramanan_12_ADL, thu-read_17, You-Do_Damen_14, Ortis2017OrganizingEV}, different daily-living scenarios \cite{Ego4D2021} and relying on different modalities such as depth \cite{thu-read_17} and gaze \cite{Li2018_EGTEA-GAZE+}.
While these contexts provide interesting test-beds to study human behavior in general, the industrial domain (e.g., factories, building sites, mechanical workshops, etc.) has never been explored from FPV. This is mainly due to the fact that data acquisition in industrial domains is difficult because of privacy issues and the need to protect industrial secrets \cite{privacy_protection_Yee}.

Nowadays many wearable glasses are able to capture different signals such as IMU, depth maps, as well as pupils fixations and audio (e.g., Microsoft Hololens2, Nreal Light, Magic Leap). Multimodal data are of relevant importance because they can be used to represent the same observed scene with complementary information. Moreover, each different signal provides additional information about the observed environment and the camera wearer, such as semantic information (RGB), 3D information of the environment and the objects (depth), as well as  the user's attention (gaze). Despite the availability of such multimodal signals in many wearable platforms available on the market, current datasets in egocentric vision seldom include rich multimodal signals.

In this paper, we present MECCANO Multimodal, which comprises multimodal egocentric data acquired in an industrial-like domain. This dataset extends the previous MECCANO egocentric video dataset \cite{ragusa2020meccano} considering two extra modalities (i.e., depth and gaze signals), a new set of annotations (i.e., temporal action annotations and spatial bounding boxes of hands and next-active objects) and a new benchmark addressing 5 different tasks aimed to understand the human behavior exploiting different signals (i.e., RGB, depth and gaze). To collect the dataset, we asked 20 subjects to build a toy model of a motorbike (see Figure~\ref{fig:toy_model}) which is composed of 49 components with different shapes and sizes. Similarly to what happens in an industrial scenario, the subjects interact with tools such as a screwdriver and a wrench, as well as with tiny objects such as screws and bolts while executing a task involving sequential actions (e.g., take wrench, tighten bolt, put down wrench). Despite the fact that this scenario is a simplification of what can be found in real industrial settings, it is still fairly complex to model, as our experiments show. The dataset has been acquired in two countries (Italy and United Kingdom) using a custom headset. The multimodality is characterized by the gaze signal, depth maps and RGB videos acquired simultaneously with two different devices (additional details are reported in Section~\ref{sec:dataset}). We acquired 20 RGB videos associated to the 20 depth videos with a Intel RealSense SR300\footnote{https://ark.intel.com/content/www/us/en/ark/products/92329/intel-realsense-camera-sr300.html}. 
In addition, we captured the gaze signal using a Pupil Core device\footnote{https://pupil-labs.com/products/core/} and synchronized it with the RGB videos. 
MECCANO has been annotated to address different tasks related to human behavior understanding. Specifically, we provide temporal annotations indicating the start and the end times of each action performed by the participants and the contact time which indicates the first frame of contact between the hand and the object, i.e., when the object changes its state from \textit{passive} to \textit{active}. We also spatially annotated the objects involved in the interactions (i.e., active objects) and the hands of the subjects with bounding boxes. Moreover, starting from the active objects annotations, we labeled the same objects in the past to explore the task of predicting the future intentions of subjects by detecting and recognizing the next active objects (i.e., the next objects the user is going to interact with). The dataset is publicy released at the following link: \url{https://iplab.dmi.unict.it/MECCANO/}. 

To highlight the usefulness of the proposed multimodal dataset, we release baseline experiments related to five fundamental tasks focused on understanding human’s behaviour from first person vision in the considered industrial-like context: 1) Action Recognition, 2) Active Objects Detection and Recognition, 3) Egocentric Human-Objects Interaction Detection, 4) Action Anticipation and 5) Next-Active Objects Detection and Recognition.
Some of these tasks have been treated in the state of the art, while, egocentric human-object interaction (EHOI) detection and next-active objects (NAO) detection and recognition tasks are underexplored considering the egocentric point of view.
We revisit these tasks considering the FPV paradigm in Section \ref{sec:EHOI} and \ref{sec:NAO} respectively.
Results demonstrate that solving these problems in the industrial domain settings from an egocentric point of view is challenging despite the availability of multimodal signals.

In sum, the contributions of this work are as follows: 1) we present MECCANO multimodal, a new challenging egocentric multimodal dataset related to the industrial domain; 2) we study in details the HOI task considering the Egocentric Vision paradigm (EHOI); 3) we study the Next-Active Object Detection task from the egocentric perspective; 4) we propose a benchmark aimed to study human behavior in the considered industrial-like scenario exploring five different tasks, showing that the current state-of-the-art approaches are not sufficient to solve the considered problems in the industrial settings.

The remainder of the paper is organized as follows. In Section~\ref{sec:related_work} we discuss related work. The proposed MECCANO Multimodal dataset is presented in Section~\ref{sec:dataset}. Section~\ref{sec:benchmark} describes the benchmark and discussed the results. We conclude the paper and discuss insights for future work in Section~\ref{sec:conclusion}.
\section{Related Work}
\label{sec:related_work}
Our work is related to different lines of research, including the collection of benchmark datasets, action recognition, HOI detection, Egocentric HOI detection, action anticipation and next-active objects detection. The following sections discuss the relevant works belonging to the aforementioned research areas.

\subsection{Datasets for Human Behavior Understanding} 
Different third person vision datasets have been proposed to study human behavior understanding exploiting the tasks of Human-Object Interaction (HOI) detection \cite{Gupta2015VisualSR, HICO_Chao, PPDM_liao2019} and action recognition \cite{caba2015activitynet, Kinetics_2017, Kinetics_Carreira2019ASN, Something_Something_Goyal}. These datasets are often composed of RGB images or videos as well as other modalities such as depth \cite{Li2010ActionRB, Wang2012MiningAE, Sung_CAD60, Koppula_cad120, Chen_UTD, Rahmani2016HistogramOO, Hu_3DHOI, Liu2020NTUR1}, especially after the release of the Microsoft Kinect \cite{zhang_kinect}. \cite{Gupta2015VisualSR} annotated the COCO dataset~\cite{lin2014COCO} with verbs (V-COCO) to study the problem of detecting HOI. V-COCO includes 10346 images annotated with 26 actions. HICO-Det \cite{HICO_Chao} is a large-scale dataset composed of static images used as a benchmark to study the task of HOI detection. This dataset includes 47766 images and has been annotated with 117 verbs and 80 objects (same objects as COCO). While these datasets focused on common and general actions, the HOI-A dataset~\cite{PPDM_liao2019} focused on a subset of actions, such as \textit{smoking cigarette} or \textit{talk on mobile phone} which can be considered dangerous actions while driving. The dataset is composed of 38668 images annotated with 10 verbs and 11 object classes.
ActivityNet \cite{caba2015activitynet} is a large-scale dataset composed of videos depicting activities that are related to how humans spend their time in their daily lives, such as \textit{walking the dog} or \textit{hand-washing clothes}. The dataset is composed of a total of 849 video hours including 203 activity classes. \cite{Kinetics_2017} and \cite{Kinetics_Carreira2019ASN}~presented Kinetics, which is a third person video dataset related to human actions. The dataset is composed of 700 human action classes which include human-object interactions, such as \textit{play instrument} and human-human interactions, such as \textit{shake hands}.
For each action, there are at least 600 video clips taken from YouTube videos.
The authors of~\cite{Something_Something_Goyal} proposed Something-Something, a video dataset which includes low-level concepts (``\textit{something-something}'') to represent simple everyday aspects of the world. It contains 108499 short videos (from 2 to 6 seconds) annotated with 174 textual description such as ``turning \textit{something} upside down'' or ``spilling \textit{something} next to \textit{something}''.

Other works have considered the egocentric scenario investigating different domains. Egocentric activities related to daily living have been studied by the authors of~\cite{ADL_PirsiavashR12}, who proposed the ADL dataset. The dataset is composed of one million frames acquired by 20 people performing a set of 18 actions of daily activities in their own apartments.
The 3D~structure of the scenes has been explored by the authors of~\cite{moghimi_ego_RGBD} who proposed a dataset composed of 5 sequences acquired using an RGB-D camera by 4 different users. The authors of~\cite{thu-read_17} proposed a video-based RGB-D egocentric dataset (THU-READ) including different types of daily-life actions. The egocentric videos have been captured in 5 scenarios such as laboratory, bathroom, conference room, dormitory and restaurant by 8 different subjects performing 40 different actions.
The problem of 3D hand-object actions recognition has been addressed by the authors of~\cite{GarciaHernando2018FirstPersonHA}, which released the Daily hand-object actions dataset containing 1175 videos belonging to 45 action categories. The dataset has been acquired by 6 actors over 3 different scenarios. A total of 105.459 RGB-D frames have been acquired and annotated with hand pose and action categories.
Some works explored the kitchen domain from the first person point of view.
Among these, the authors of~\cite{Torre2009CMU-MMAC} released the CMU Multi-Modal Activity Database (CMU-MMAC) to study human activities in a kitchen environment. The authors built a kitchen and acquired egocentric videos from 5 different subjects cooking 5 recipes. They captured RGB videos using different cameras, audio and motion capture information.
EPIC-Kitchens and its extension \cite{Damen2018EPICKITCHENS, Damen2020Collection, Damen2020RESCALING} are a series of egocentric datasets focused on unscripted activites related to kitchens. In particular, EPIC-Kitchens-55 \cite{Damen2018EPICKITCHENS} is composed of 432 videos annotated with 352 objects classes and 125 different verbs classes. EPIC-Kitchens-100 \cite{Damen2020Collection} is an extension of EPIC-Kitchens-55 in terms of videos (700), environments (45) and hours (100). Along with the dataset, the authors proposed 6 challenges to study human behavior understanding in kitchens: action recognition, action detection, action anticipation, domain adapatation for action recognition, object detection and multi-instance retrieval. The authors of~\cite{Li2018_EGTEA-GAZE+} studied egocentric video action recognition, considering both RGB and gaze signals to determine what a person is doing (action recognition) and where they are looking (gaze estimation). They presented the dataset EGTEA Gaze+, where 32 subjects perform 7 different meal preparation tasks in different kitchens. EGTEA Gaze+ is composed of 106 action classes and includes gaze information collected at every frame.  \\
Not only depth and gaze signals have been considered in past works. Sensor data like accelerometer or gyroscope have also been considered to recognize egocentric activity. The authors of~\cite{Song_sensor_data} captured a dataset of egocentric videos using a Google Glass, that acquired RGB videos and sensor information. In particular, 200 short sequences have been acquired by 20 different subjects which performed daily activities.
\cite{Rogez_object_grasp} focused on object manipulation and proposed the Grasp UNderstanding (GUN-71) dataset which is composed of 12.000 RGB-D images labeled with 71 grasp classes. The videos have been acquired by 8 different subjects which performed different grasps on personal objects in 5 different houses. The camera used is a chest-mounted Intel'Senz3D\footnote{https://it.creative.com/p/archived-products/blasterx-senz3d} which is a webcam paired with a depth sensor.
Beyond the aforementioned datasets which considered only one extra modality in addition to the RGB signal, the authors of~\cite{Kothari2020GazeinwildAD} proposed the Gaze-in-Wild dataset which comprises both gaze and depth streams. This dataset has been acquired by 19 participants which performed 4 activities: indoor navigation, ball catching, object search and tea making. They used a Pupil Labs eye tracker to acquire the gaze signal, the MPU to obtain the IMU data and the ZED stereo camera to acquire the depth map. Recently, Ego4D, a massive-scale egocentric video dataset has been released \cite{Ego4D2021}. Is has been acquired by 931 camera wearers from 9 different countries. Ego4D comprises videos, audio, 3D meshes of the environment, eye gaze, stereo and videos acquired by multiple egocentric cameras. The data has been collected in multiple domains and comprises different activities such as people playing cards, working at desk, cleaning the garden, cooking something or practicing a musical instrument. In addition to the dataset, \cite{Ego4D2021} presented five benchmarks focused on egocentric perception in the past, present and future.

Inspired by the first version of the MECCANO dataset \cite{ragusa2020meccano}, \cite{Sener2022Assembly101AL} proposed Assembly101 which is a procedural activity dataset comprising multi-view videos in which subjects assembly and disassembly toys. Contextually, they benchmarked three action understanding tasks (i.e., action recognition, action anticipation and temporal segmentation) and proposed a new task which is related to mistakes detection. Despite the similar setup to MECCANO, we focus on the multimodal nature of the acquired data. Moreover, they acquired egocentric videos with monochrome cameras using a device similar to Oculus Quest VR. In addition, Assembly101 is able to address tasks related only to the actions performed by the users.
 
It is worth noting that previous egocentric datasets have considered scenarios related to kitchens, offices, and daily-life activities and that they have generally tackled the action recognition task rather than EHOI detection.
Table~\ref{tab:datasets} compares the aforementioned datasets with respect to the proposed MECCANO Multimodal dataset which is a conspicuous extension of the previous MECCANO dataset \cite{ragusa2020meccano}.

As shown in Table~\ref{tab:datasets}, MECCANO Multimodal is the first egocentric multimodal dataset comprising both gaze and depth signals acquired in an industrial-like domain. Moreover, it has been explicitly annotated to tackle different tasks with different modalities useful to build a real system able to support humans in the industrial domain: 1) Action Recognition, 2) Active Object Detection and Recognition, 3) Egocentric Human-Object Interaction, 4) Action Anticipation and 5) Next-Active Object Detection.

\begin{table*}[]
\caption{Comparison of MECCANO with other datasets. AA: Action Anticipation. AOD: Active Object Detection. AOR: Active Object Recognition. AR: Action Recognition. AVD: Audio-Video Diarization. AVL: Audio-Video Localization. AVT: Audio-Video Transcription. DA-AR: Domain Adaptation for Action Recognition. EHOI: EHOI Detection. FHP: Future Hand Prediction. HOI: HOI Detection. HPE: Hand Pose Estimation. LAM: Looking-at-Me. LOC: Localization. MD: Mistake Detection. MQ: Moment Queries. MR: Multi-Instance Retrieval. NAO: Next-Active Objects Detection. NLQ: Natural Language Queries. OD: Object Detection. OSCC: Object State Change Classification. PNRTL: Point-of-No-return Temporal Localization. SCOD: State change Object detection. S\&LTA: Short and Long Term Anticipation. TAS: Temporal Action Segmentation. TTM: Talking-to-Me. VQ2D\&3D: Visual Queries with 2D\&3D Localization.}
\label{tab:datasets}
\resizebox{\textwidth}{!}{%
\setlength\tabcolsep{2pt}
\begin{threeparttable}
\begin{tabular}{lcccllcccccccc}
\multicolumn{1}{c}{\textbf{Dataset}} & \textbf{Settings} & \textbf{EGO?} & \textbf{Video?} & \multicolumn{1}{c}{\textbf{Signals}} & \multicolumn{1}{c}{\textbf{Tasks}} & \textbf{Year} & \textbf{Frames} & \textbf{Sequences} & \textbf{AVG. video duration} & \textbf{Action classes} & \textbf{Object classes} & \textbf{Object BBs} & \textbf{Participants} \\ \hline
MECCANO Multimodal & Industrial-like & \checkmark & \checkmark & RGB, depth, gaze & \begin{tabular}[c]{@{}l@{}}EHOI, AR, AOD, AOR,\\ AA, NAO\end{tabular} & 2022 & 299,376 & 20 & 20.79 min & 61 & 20 & 307,601 & 20 \\ \hline
Assembly101 \cite{Sener2022Assembly101AL} & Industrial-like & \checkmark & \checkmark & RGB, multi-view, 3D hand-pose & AR, AA, TAS, MS & 2022 & 20M & 362 & 7.10 min & 1380 & 90 & 0 & 53 \\ 
EGO4D\tnote{1} \cite{Ego4D2021} & Multi Domain & \checkmark & \checkmark &\begin{tabular}[c]{@{}l@{}}RGB, Audio, 3D environments, \\ stereo, gaze, IMU, multi-view\end{tabular}  &\begin{tabular}[c]{@{}l@{}}VQ2D\&3D, NLQ, MQ, PNRTL, \\ SCOD, OSCC, AVD, AVT, AVL,\\ LAM, TTM, S\&LTA, FHP\end{tabular}  & 2022 & 418M\tnote{2}  & 9650 & 24.11 min & 113\tnote{3} & 449\tnote{3} & 295104\tnote{4}& 931 \\ 
EPIC-KITCHENS-100 \cite{Damen2020RESCALING} & Kitchens & \checkmark & \checkmark & RGB & AR, AD, AA, DA-AR, MR & 2021 & 20M & 700 & N/A & 97 & 300 & N/A & 37 \\ 
Gaze-in-wild \cite{Kothari2020GazeinwildAD} & Daily activities  & \checkmark & \checkmark & RGB, depth, gaze & AR & 2020 & N/A & N/A & N/A & 4 & 0 & 0 & 19 \\
EGTEA Gaze+ \cite{Li2018_EGTEA-GAZE+} & Kitchens & \checkmark & \checkmark & RGB, gaze & AR & 2018 & 2,4M & 86 & 0.05 min & 106 & 0 & 0 & 32 \\
Daily Hand-Object Actions \cite{GarciaHernando2018FirstPersonHA} & Daily activities & \checkmark & \checkmark & RGB, depth & AR, HPE & 2018 & 105,459 & 1175 & 0.05 min & 45 & 26 & N/A & 6 \\
THU-READ \cite{thu-read_17}& Daily activties & \checkmark & \checkmark & RGB, depth & AR & 2017 & 343,626 & 1920 & 7.44 min & 40 & 0 & 0 & 8 \\
Multimodal Egocentric Activity \cite{Song_sensor_data} & Daily activities & \checkmark & \checkmark & RGB, sensor data & AR & 2016 & 30,000 & 200 & 0.25 min & 20 & 0 & 0 & 20 \\
GUN-71 \cite{Rogez_object_grasp} & Daily activities & \checkmark & \checkmark & RGB, depth & AR & 2015 & 12,000 & N/A & N/A & 71 & 28 & 0 & 8 \\
Wearable Computer Vision System \cite{moghimi_ego_RGBD} & Daily activities & \checkmark & \checkmark & RGB, depth & AR & 2014 & N/A & 5 & N/A & 12 & 0 & 0 & 4 \\ 
ADL \cite{ADL_PirsiavashR12} & Daily activities & \checkmark & \checkmark & RGB & AR, AOR & 2012 & 1,0M & 20 & 30.0 min & 32 & 42 & 137,780 & 20 \\ 
CMU \cite{Torre2009CMU-MMAC} & Kitchens & \checkmark & \checkmark & RGB & AR & 2009 & 200,000 & 16 & 15.0 min & 31 & 0 & 0 & 16 \\ 
\hline
NTU RGB+D 120 \cite{Liu2020NTUR1} & General & X & \checkmark & RGB, depth & AR & 2020 & 8 M & 114,480 & N/A & 120 & 0 & 0 & 106 \\
UTD-MHAD \cite{Chen_UTD} & General & X & \checkmark & RGB, depth, sensor data & AR & 2017 & N/A & 861 & N/A & 27 & 0 & 0 & 8 \\
SYSU 3D Human-Object Interaction \cite{Hu_3DHOI} & General & X & \checkmark & RGB, depth & AR & 2017 & N/A & 480 & N/A & 12 & 0 & 0 & 40 \\
Something-Something \cite{Something_Something_Goyal} & General & X & \checkmark & RGB & AR, HOI & 2017 & 5,2 M & 108,499 & 0.07 min & 174 & N/A & 318,572 & N/A \\
Kinetics \cite{Kinetics_2017} & General & X & \checkmark & RGB & AR & 2017 & N/A & 455,000 & 0.17 min & 700 & 0 & 0 & N/A \\
UWA3D Multiview Activity \cite{Rahmani2016HistogramOO}& General & X & \checkmark & RGB, depth & AR & 2016 & N/A & 1200 & N/A & 30 & 0 & 0 & 10 \\
ActivityNet \cite{caba2015activitynet} & Daily activities & X & \checkmark & RGB & AR & 2015 & 91,6 M & 19,994 & 2.55 min & 200 & N/A & N/A & N/A \\
CAD-120 \cite{Koppula_cad120} & General & X & \checkmark & RGB, depth & AR, AOR & 2013 & 61585 & 120 & 0.28 min & 10 & 12 & N/A & 4 \\
MSRDailyActivity3D \cite{Wang2012MiningAE} & Daily activites & X & \checkmark & RGB, depth & AR & 2012 & N/A & 320 & N/A & 16 & 0 & 0 & N/A \\
Human Activity Detection \cite{Sung_CAD60} & Daily activites & X & \checkmark & RGB, depth & AR & 2011 & N/A & N/A & 0.75 min & 12 & 0 & 0 & 4 \\
MSR-Action3D \cite{Li2010ActionRB} & General & X & \checkmark & depth & AR & 2010 & 23797 & 402 & 0.07 min & 20 & 0 & 0 & 7
\end{tabular}
\begin{tablenotes}
    \item[1] The statistics have been obtained on May 15, 2022.
    \item[2] The number of frames has been obtained considering the canonical videos acquired at 30 fps.
    \item[3] This number has been obtained from the ``Long-Term Anticipation" task considering both Training and Validation sets.
    \item[4] The number of bounding boxes has been obtained considering the ``State Change Object Detection" task.
  \end{tablenotes}
\end{threeparttable}
}
\end{table*}

\subsection{Human Behavior Understanding Tasks}
In this section we discuss the state of the art focusing on relevant tasks which should be exploited to understand the human behavior.
\subsubsection{Action Recognition}
Video action recognition has been thoroughly studied by researchers, especially from the third person view.
Some works \cite{Learning_actions_movies_Laptev, Human_detection_Flow_Schmid_06, TwoStream_convolutional_action_Zisserman_14, Two-Stream_Zisserman, temporal_segNet} mixed classic approaches considering hand-crafted features, such as optical flow and deep networks to represent the motion of actions using two stream networks. 
3D ConvNets are commonly used to encode both spatial and temporal dimensions in a unified way \cite{ Conv_spatio-temporal_Taylor, Learning_spatio-temporal_Paluri, Carreira2017QuoVA}. Long-term filtering and pooling has focused on representing actions considering their full temporal extent \cite{Long-term_action_Schmid, Two-Stream_Zisserman, temporal_segNet, Zhou2018TemporalRR}. 
Other works separately control spatial and temporal dimensions factoring convolutions into separate 2D spatial and 1D temporal filters \cite{Spatiotemporal_residual_action, closer_spatiotemp_action, rethinking_spatiotemporal, Learning_spatiotemporal_pseudo}.
Slow-Fast networks \cite{feichtenhofer2018slowfast} avoid using pre-computed optical flow and encodes the motion of actions into a ``fast'' pathway (which operates at a high frame rate) and simultaneously a ``slow'' pathway which captures semantics (operating at a low frame rate). 
The authors of~\cite{Zhou2018TemporalRR} introduced a network module called Temporal Relation Network (TRN) to learn temporal relations between video frames at multiple time scales. The authors of~\cite{TSM_2019} proposed a temporal shift module (TSM). This module allows 2D architectures to obtain comparable performance to 3D CNNs.
Inspired by feature selection methods, the authors of~\cite{Feichtenhofer2020X3DEA} presented a family of video networks (X3D) which expand a 2D image classification architecture into a spatiotemporal one by expanding along multiple possible axes such as space, time, width and depth.
The authors of~\cite{Hussein2019TimeceptionFC} revisited the definition of \textit{activity} and restricted it to \textit{complex action} which is a set of simple one-actions which compose the activity (e.g., \textit{cooking a meal} can be considered as a set of one-actions: \textit{get, cook, put and wash}). Recently, the action recognition task has been addressed considering multi-modal signals. For example, the authors of~\cite{Shi_2021_ICCV} considered audio, visual and textual information to recognize actions using graph convolutional neural networks (GCN).
Previous works also investigated egocentric action recognition adapting third person vision approaches to the first person scenario \cite{TSM_2019, Zhou2018TemporalRR, feichtenhofer2018slowfast, Damen2018EPICKITCHENS}.

In this work, we assess the performance of state-of-the-art action recognition methods on the proposed MECCANO dataset considering the state-of-the-art SlowFast network \cite{feichtenhofer2018slowfast} as a baseline.

\subsubsection{HOI Detection}
Previous works have investigated HOI detection mainly from a third person vision point of view. The authors of~\cite{Gupta2015VisualSR} were the first to explore the HOI detection task annotating the COCO dataset \cite{lin2014COCO} with verbs. The authors proposed a method to detect people performing actions able to localize the objects involved in the interactions on still images. The authors of~\cite{Gkioxari2018DetectingAR} proposed a human-centric approach based on a three-branch architecture (InteractNet) instantiated according to the classic definition of HOI in terms of a $<$human, verb, object$>$ triplet. This approach analyzes each human-object pairs detected with an object detector~\cite{ren2015faster} using a heat map to represent their relationship.  Some works~\cite{Qi2018LearningHI, Chao2018LearningTD, RPN_Zhou} explored HOI detection using graph convolutional neural networks after detecting humans and objects in the scene. Recent works \cite{PPDM_liao2019, Wang_InteractionPoints_2020_CVPR} represented the relationship between both, the humans and the objects, as the intermediate point which connects the center of the human and object bounding boxes.
The aforementioned works addressed the problem of HOI detection in the third person vision domain. In this work, we look at the task of HOI detection from an egocentric perspective considering the proposed MECCANO Multimodal dataset.

\subsubsection{Tasks related to EHOI Detection}
The problem of Human-Object Interaction (HOI) detection has been systematically investigated only from Third Person Vision.
Previous work have considered similar tasks related to Egocentric Human-Object Interaction (EHOI) detection due to the limited availability of egocentric datasets explicitly labelled for this task. 
Some studies have modeled the relations between entities for interaction recognition as object affordances~\cite{Hotspots_Grauman19, Nagarajan2020EGOTOPOEA, affordance_Fang18}. Other studies tackled tasks related to EHOI recognition proposing hand-centric methods \cite{Cai2016UnderstandingHM, Lending_Hand_Bambach_15, Hands_in_contact_Shan20, kwon2021h}. The authors of~\cite{Hands_in_contact_Shan20} proposed to detect and localize hands in the scene distinguishing left from right hands. Objects are classified into two classes: \textit{active} or \textit{passive}. In particular, if an object is in contact with at least one hand, it is considered as \textit{active}, otherwise, it is considered as \textit{passive}. The authors of~\cite{Li_Adaptive_2020_CVPR} proposed to search network structures with differentiable architectures to construct adaptive structures for different videos to facilitate adaptive interaction modeling. The method has been evaluated on the Something-Something dataset \cite{Something_Something_Goyal} which contains egocentric-like videos. The authors of~\cite{kwon2021h} proposed a unified approach to recognize hand-object interactions predicting the 3D pose of the two interacting hands and the 6D pose of manipulated objects simultaneously.
Despite these works have considered tasks related to human-object interaction from an egocentric point of view, the EHOI detection task has not yet been studied systematically. In this work, we formalize the task of EHOI detection and focus on this problem related to the industrial domain, considering the proposed MECCANO dataset.

\begin{figure*}[t]
\centering
\includegraphics[width=\textwidth]{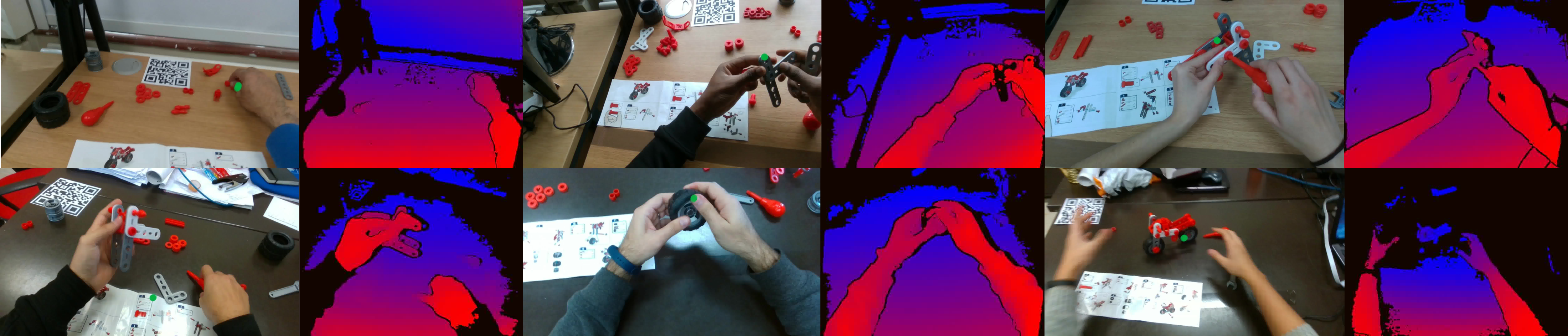}
\caption{Examples of data acquired by the 20 different participants in two countries (Italy, United Kingdom).}
\label{fig:dataset}
\end{figure*}

\subsubsection{Action Anticipation}
The task of action anticipation has been investigated from the third person point of view \cite{Hierarchical_Repr_Savarese_14, Felsen_what_will_happen_17, Gao2017REDRE, Zeng2017VisualFB}. \cite{Hierarchical_Repr_Savarese_14} proposed a new representation of actions called \textit{hierarchical movemes} to anticipate the future actions from still image or short video clips. They encode the atomic components of human movements before an action is executed, representing the actions with high semantic and temporal granularity. The authors of~\cite{Felsen_what_will_happen_17} proposed a generic framework for forecasting future events in team sports videos related to water polo and basketball events. The authors of~\cite{Gao2017REDRE} considered multiple history representations of the past to anticipate a sequence of future representations.
In the last years, the task has been studied also from the first person perspective. 
The authors of~\cite{robot-centric-anticipation_15} explored this task with the aim to assist humans which cooperate with a robot. In particular, they anticipate future actions considering videos acquired from the point of view of a robot which interacts with a human. A series of works \cite{furnari2019rulstm, furnari2020rulstm, slowfast_rulstm_ballan} address the task considering the LSTM networks to encode the features related to the past. The authors of~\cite{Roy_2022_WACV} focused on the goal representation to predict the next action from the first person view.

In this work, we adopt the RULSTM model \cite{furnari2020rulstm} to evaluate state-of-the-art action anticipation methods on the proposed MECCANO dataset also considering the gaze and depth signals.

\subsubsection{Next Active Objects Detection}
The detection of next-active objects, i.e., the objects which will be involved in a EHOI, is a problem which has not been thoroughly studied due to the small number of public egocentric datasets suitable for the task.
There are not egocentric datasets specifically annotated to perform this task. 
The authors of~\cite{Furnari2017NextactiveobjectPF} have been the first to explore the next-active objects prediction problem. They performed experiments on the Activity of Daily Living (ADL) egocentric dataset, analyzing the trajectories of the next-active objects with a temporal sliding window. 
The authors of~\cite{liu_forecasting_HOI} addressed the task of anticipating egocentric actions proposing an architecture composed of a motor attention module to predict the trajectory of the hands and by a module to detect the area of contact of the target object which will be active. These two outputs are fed into an anticipation module which predicts a spatio-temporal attention map which indicates the possible location of the next-active objects. The output is composed of the next action label, a ``hotspot" which indicates the area of the object in which there will be a contact, and the hand trajectory.
The two egocentric datasets ADL \cite{ADL_PirsiavashR12} and EPIC-Kitchens \cite{Damen2020RESCALING} have been re-annotaded by \cite{JIANG2021212} to tackle the problem of short-term next-active object detection. They proposed a novel human-centerd approach composed of two pathways: 1) the first pathway generates a human visual attention probability map and 2) the second one generates a human hand position probability map. These two maps are then fused by an interaction module which outputs the final map of the next-active object.
In addition to the next-object location, \cite{Fan2017ForecastingHA} predicted the hands location in future frames. The problem was tackled designing a two-stream CNN architecture with an
auto-encoder by extending SSD, a state-of-the-art convolutional object detection network and using a regression network to infer future representations. 
The authors of~\cite{Dessalene_forecasting_contact} performed action anticipation and prediction through hand-objects contact representations. They presented a new architecture composed of an anticipation module and temporal relations represented using a Graph Convolutional Networks (GCN) and the LSTMs to predict the final next action label. They treated the next-active objects involved into the future contact with hands through semantic segmentation masks. 
The authors of~\cite{Bertasius2017FirstPA} detected the important objects for the camera wearer (i.e., objects related to the intent of the user) considering an unsupervised learning approach. Closely related to next-active object prediction, the authors of~\cite{Ego4D2021} presented a set of tasks for the forecasting benchmark of the Ego4D dataset including the task of short-term Object Interaction Anticipation.

Despite previous works have considered tasks related to the problem of next-active object detection, it has not yet been studied in depth.
Moreover, the task has not been studied considering different types of signals (e.g., depth and gaze) and in an industrial-like domain.
In this work, we focus on the next-active object task on the MECCANO Multimodal dataset, which has been acquired in an industrial domain and annotated explicitly to tackle this challenging task.
\section{The MECCANO Multimodal Dataset}
\label{sec:dataset}
In this Section, we describe MECCANO multimodal, a dataset of egocentric videos composed of multimodal data collected in an industrial-like domain. We acquired RGB videos, depth maps and gaze signal simultaneously with two different devices.

\subsection{Data Collection}
\label{sec:meccano_collection}
\begin{figure}[t]
	\centering
	\includegraphics[width=\columnwidth]{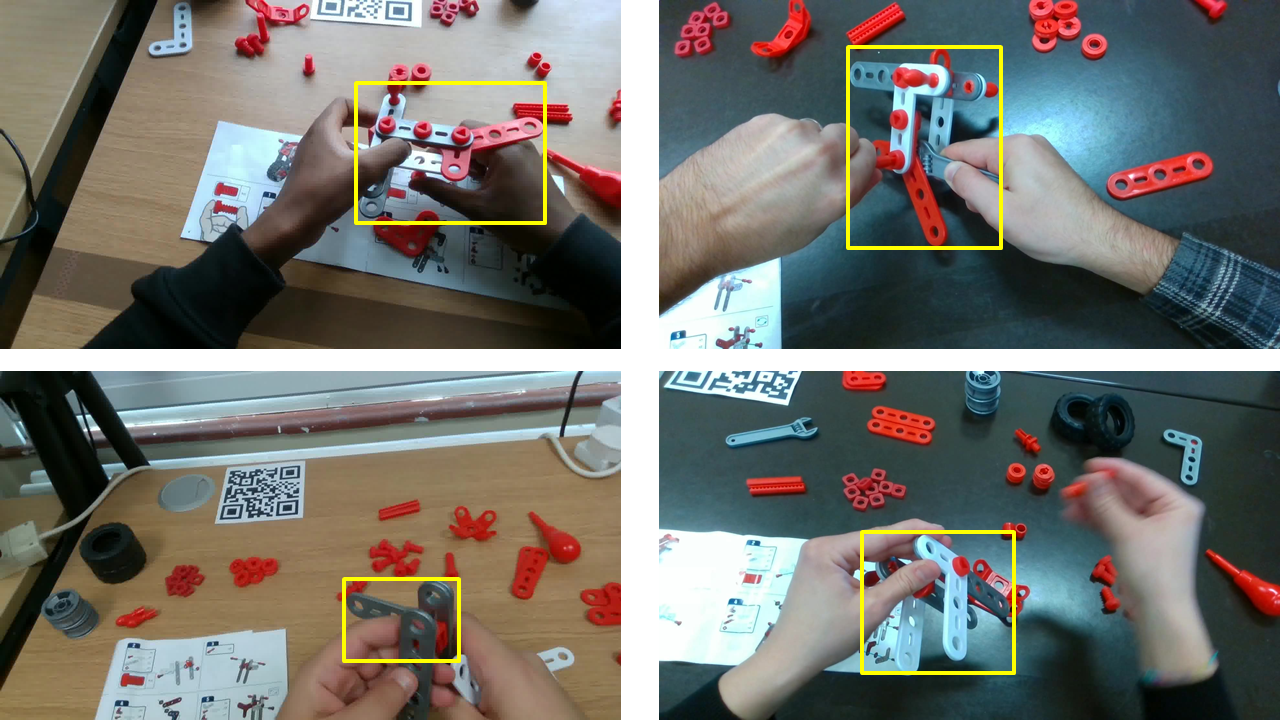}
	\caption{Examples of objects belonging to the partial model class.}
	\label{fig:partial}
\end{figure}

The MECCANO multimodal dataset has been acquired in an industrial-like scenario in which 20 subjects were asked to built a toy model of a motorbike (see Figure~\ref{fig:toy_model}). The motorbike is composed of 49 components with different shapes and sizes belonging to 19 classes. In addition, 2 tools, a \textit{screwdriver} and a \textit{wrench}, are available to facilitate the assembly of the toy model. In our settings, we have grouped two types of components which are similar in their appearance and have similar roles in the assembly process. Specifically, we grouped the A054 and A051 components (shown in Figure~\ref{fig:toy_model}) under the ``screw'' class. These two types of components only differ in their lengths. We also grouped A053, A057 and A077 under the ``washers'' class. Note that these components only differ in the radius of their holes and in their thickness. As a result, we have 20 object classes in total: 16 classes are related to the 49 motorbike components, whereas the others are associated to the two tools, to the instruction booklet and to a ``partial model" class, which indicates a set of components assembled together to form a part of the model (see Figure~\ref{fig:partial}). Note that multiple instances of each component are necessary to build the model.

\begin{figure}[t]
\centering
\includegraphics[width=\columnwidth]{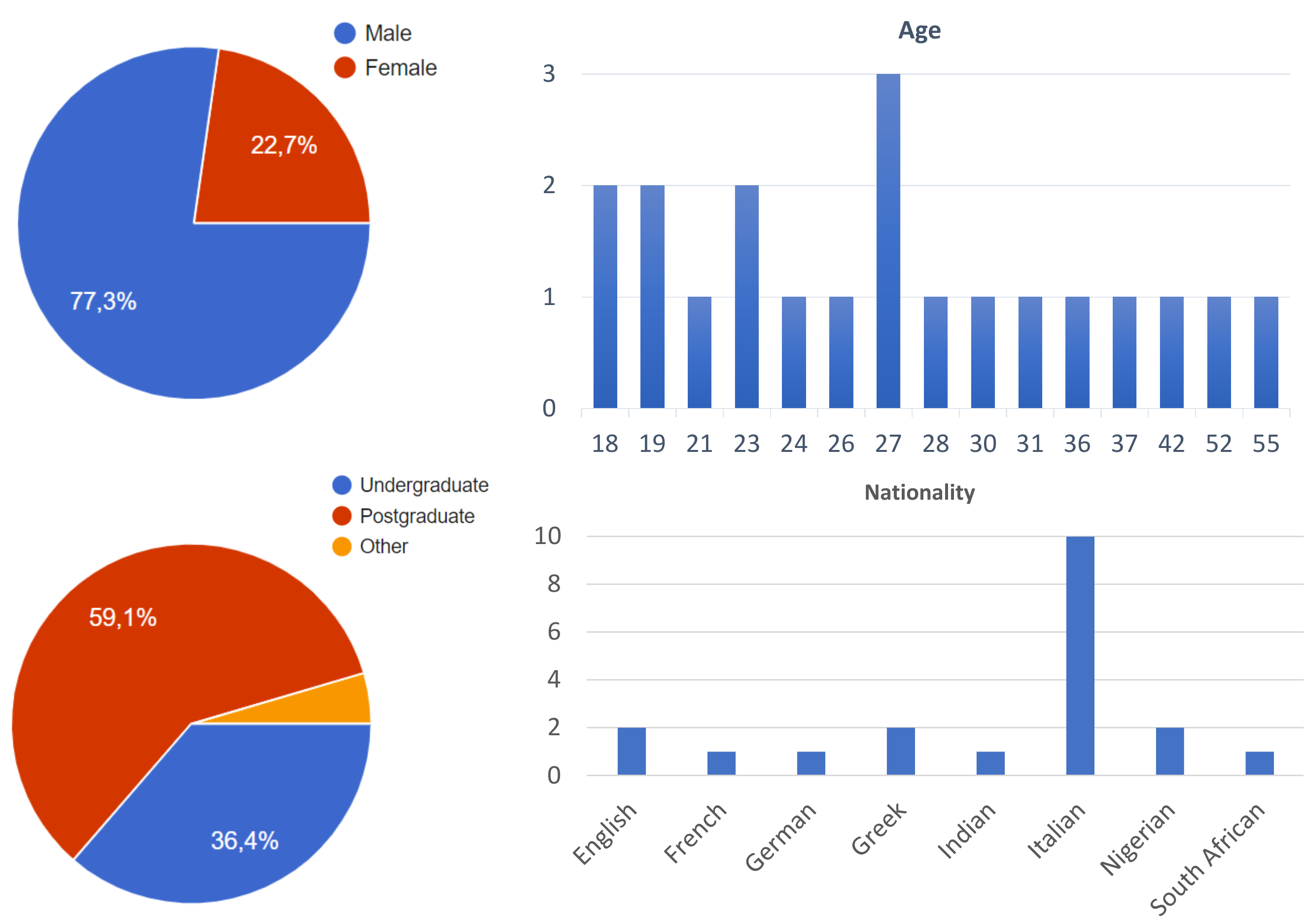}
\caption{Statistics of the 20 participants.}
\label{fig:participants}
\end{figure}

For the data collection, the $49$ components related to the $16$ considered classes, the $2$ tools and the instruction booklet have been placed on a table to simulate an industrial-like environment. Specifically, objects of the same component class have been grouped and placed in a heap, and heaps have been placed randomly on the table (see Figure~\ref{fig:dataset}). 
We have considered two types of tables: a light-colored table and a dark one. The dataset has been acquired in 2 different countries, Italy and United Kingdom. Participants were from $8$ different nationalities with ages between $18$ and $55$. Figure~\ref{fig:participants} reports some statistics about the participants. We asked participants to sit and build the model of the motorbike. No other particular instruction was given to the participants, who were free to use all the objects placed on the table as well as the instruction booklet. Some examples of the captured data are reported in Figure~\ref{fig:dataset}.

\begin{figure}[t]
\centering
\includegraphics[width=\columnwidth]{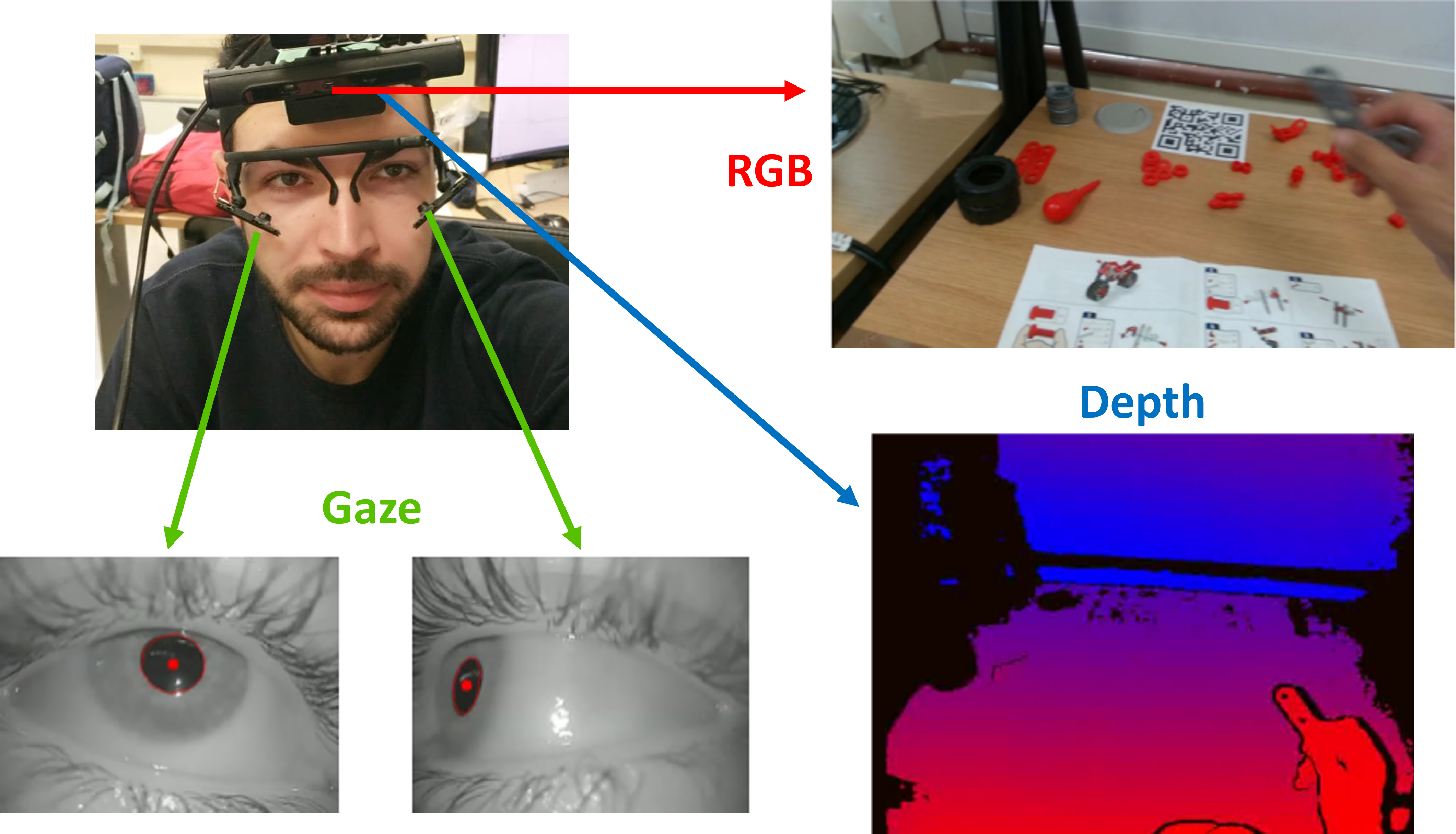}
\caption{The custom headset used to acquire the MECCANO dataset along with examples of the captured modalities. The headset is composed of two devices: a Intel RealSense SR300 and a Pupil Core.}
\label{fig:headset}
\end{figure}

The dataset has been acquired using a custom headset (see Figure~\ref{fig:headset}) which was worn by participants for acquisition purposes. The headset was composed of an Intel RealSense SR300\footnote{https://ark.intel.com/content/www/it/it/ark/products/92329/intel-realsense-camera-sr300.html} and by a Pupils Core\footnote{https://pupil-labs.com/} device. The headset was adjusted to control the point of view of the camera with respect to the different heights and postures of the participants in order to have the hands located approximately in the middle of the scene during the object manipulations. For each participant, we acquired the RGB stream and the depth signal from the RealSense device, whereas the gaze signal was acquired through the Pupils Core device (see Figure~\ref{fig:headset}). The RGB videos acquired with the RealSense device were recorded at a resolution of 1920x1080 pixels. Depth videos were acquired with a resolution of 640x480 pixels. Both videos have a framerate of 12fps. Finally, we acquired the gaze signal with the Pupils Core device with a frequency of 200Hz. To acquire the Real Sense and Pupils Core signals, we used the Pupils Capture software\footnote{https://pupil-labs.com/products/core/} which allows to acquire gaze simultaneously with signals coming from the two devices. Each video includes a complete assembly of the toy model starting from the 49 pieces placed on the table. The average duration of the captured videos is 21.14\textit{min}, with the longest one being 35.45\textit{min} and the shortest one being~9.23\textit{min}. 

\begin{figure*}[t]
	\centering
	\includegraphics[width=\textwidth]{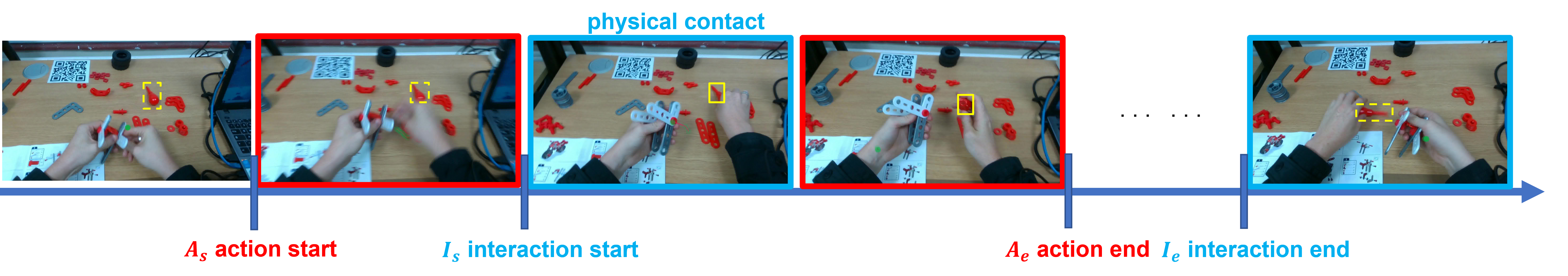}
	\caption{Example of relation between the action \textit{``take screwdriver"} and its related interaction. $A_s$ and $A_e$ represent the start and the end times of the action, while $I_S$ represents the start of the interaction represented by the physical contact between the hand and the object. The interaction will end when there will be no more physical contact ($I_e$).}
	\label{fig:action_interaction}
\end{figure*}

\subsection{Data Alignment}
We aligned the different signals both temporally and spatially to obtain a consistent association between modalities. In this way it is possible to use the set of annotations independently from the different chosen signal (e.g., temporal segments or bounding box annotations). The following sections detail the alignment of the different modalities to the source RGB videos.

\subsubsection{Depth Alignment}
There was a constant temporal misalignment of 0.4s between the depth and RGB signals due to the fact that the streams have been acquired with two different sensors (depth sensor and RGB sensor).
We temporally aligned the two streams obtaining a total of 301016 depth frames\footnote{Due to the temporal misalignment, the number of depth frames is different respect to the number of RGB frames.}. Examples of RGB frames associated with the depth maps are shown in Figure~\ref{fig:dataset}.

\subsubsection{Gaze Alignment}
The gaze data consists of 2D pixel coordinates (\textit{x, y}) of the gaze position in the RealSense RGB frame, and include also a confidence scores and the timestamps.
For each RGB frame, we associated a gaze signal selecting only gaze positions with a confidence larger than or equal to 0.6 and considering the timestamp closest to the considered frame (see Figure~\ref{fig:dataset}).

\subsection{Data Annotation}
\label{sec:meccano_data_ann}
The MECCANO Multimodal dataset has been collected and annotated to study human behavioral understanding in an industrial-like scenario.
Similarly to recent datasets \cite{Ego4D2021, Damen2018EPICKITCHENS}, MECCANO is a ``multi-task" dataset, due to its rich set of annotations which can be used and combined to solve different tasks.

We provide temporal annotations for actions and interactions understanding that are useful to solve tasks which take into account the temporal dimension, such as action recognition, as we reported in Section~\ref{sec:action_recognition}. 
Active objects have been labeled with bounding boxes and object classes with the aim to solve tasks like Active Object Detection and Recognition (Section~\ref{sec:AODR}).
Combining interaction temporal annotations and active objects bounding boxes, it is possible to address the task of Egocentric Human-Object Interaction Detection (EHOI) as detailed in Section~\ref{sec:EHOI}.
We provide also bounding boxes around the hands of the user during all object interactions. Hands and active objects bounding boxes have also been tracked backward in time before the beginning of each interaction. These tracked bounding boxes are related to the task of “next active objects”. We exploit the hands and next-active objects bounding boxes to address the action anticipation task as reported in Section~\ref{sec:Action_anticipation}. Moreover, bounding boxes of active and next-active objects have been used to solve the Next-Active Object Detection task (see Section~\ref{sec:NAO}).

In sum, to show the potential of the MECCANO Dataset and its annotations, we propose a benchmark which comprises five tasks using different sets of annotations and input modalities : 1) Action Recognition, 2) Active Object Detection and Recognition, 3) Egocentric Human-Object Interaction Detection, 4) Action Anticipation and 5) Next-Active Object Detection.
Central to human-object annotations are user-object interactions and actions. Hence, we first investigate the relationship between these two concepts.

\begin{figure}[t]
	\centering
	\includegraphics[width=\columnwidth]{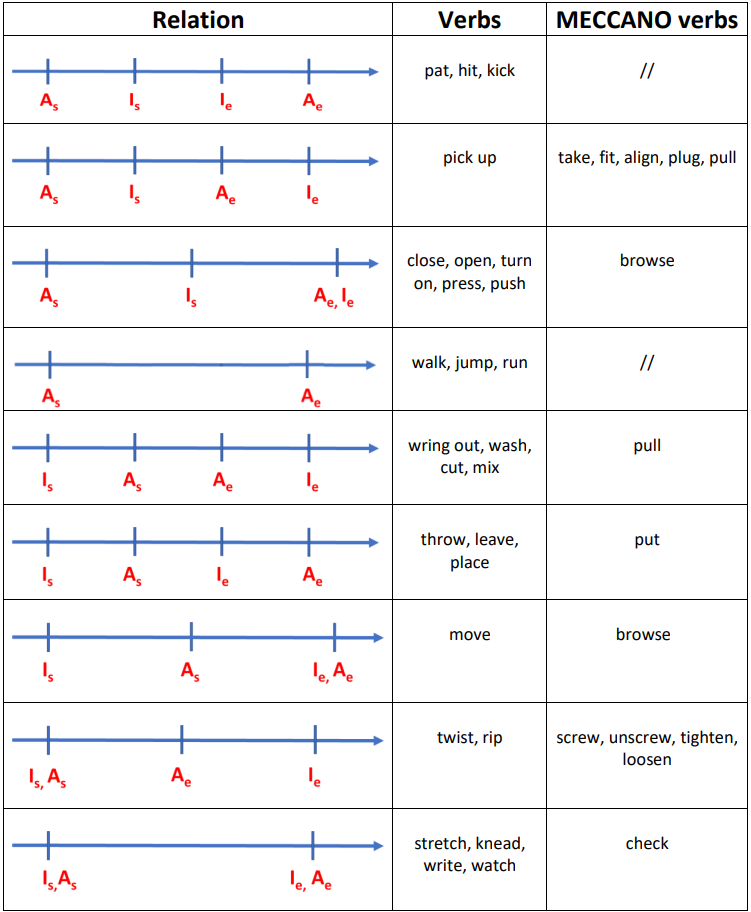}
	\caption{Examples of relations between action and interaction concepts.}
	\label{fig:action_interaction_table}
\end{figure}

\subsubsection{Action-Interaction Relations}
\label{sec:action_vs_interactions}
In the literature, the \textit{action} and \textit{interaction} concepts are often used interchangeably, specifically when tasks related to \textit{action} understanding are considered. 

\begin{figure*}[t]
\centering
\includegraphics[width=\textwidth]{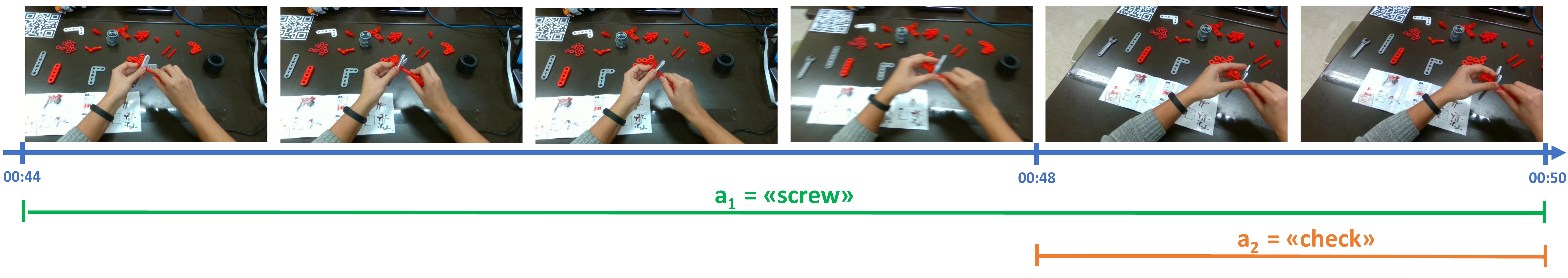}
\caption{Example of two overlapping temporal annotations along with the associated verbs.}
\label{fig:temporal}
\end{figure*}

Let's consider the sentence ``take a screwdriver'' which is composed of the verb ``take'' and the object ``screwdriver''. We consider that the \textit{interaction} is strongly related to the physical contact between the human and the object, whereas we assume that the \textit{action} is more related to the motion of the hands of the user and the related objects as well as to the change of state of the object (i.e., from inert to in hand). 
These two entities are correlated along the temporal dimension having different start and end times, also if they overlap (see Figure~\ref{fig:action_interaction}). Let $A_s$ and $A_e$ be the start and end time of an action and let $I_s$ and $I_e$ denote when the related interaction begins and ends. Considering the annotation ``take a screwdriver'', the action begins when the hand of the human starts to move towards the target object, which is the screwdriver on the table. The interaction begins when the hand touches the target object. Hence the interaction begins due to the physical contact between the hand and the object, when the action is still on-going. When the screwdriver has been taken, which means that it changed its state from inert to in hand, the action ends, while the interaction will be concluded when the physical contact will be broken (e.g., when the human will put down the object on the table).
The verb describing the action is the same which describes the related interaction. Note that the interaction can be associated to different verbs over time. For example, if after taking the screwdriver the physical contact is not broken and the human puts down the screwdriver, the verb ``put down'' will describe also the interaction until the end of the action. 

Distinguishing these two concepts and understanding their temporal correlation is fundamental to formalize and disambiguate these two related tasks. Action understanding tasks focus on the temporal dimension of actions while the Human-Object Interaction detection task focuses more on the spatial position of the objects in the scene  related to the physical contact between human and objects.

\begin{figure*}[t]
	\centering
	\includegraphics[width=\textwidth]{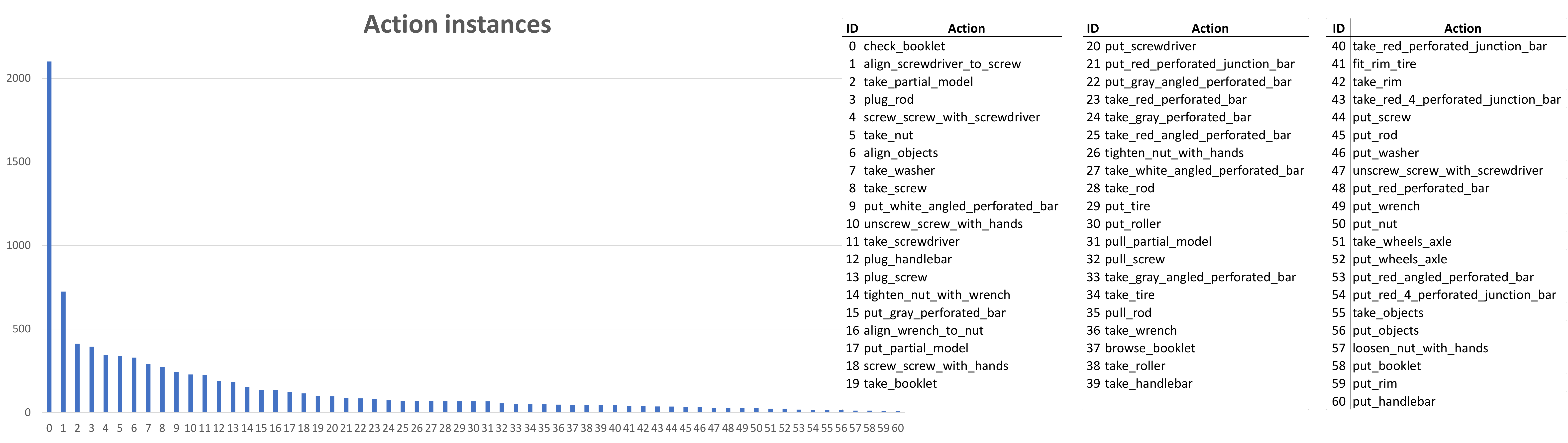}
	\caption{List of the actions present in the MECCANO Dataset (right) and their distribution (left). The action labels follows a long-taile distribution which highlights the complexity of the considered scenario.}
	\label{fig:action_stats}
\end{figure*}

Considering this distinction, we labeled the MECCANO dataset with actions and interactions annotations as reported in Section~\ref{sec:temporal_ann} and explored the two different tasks.
Figure~\ref{fig:action_interaction_table} reports different examples of relations between action and interaction concepts. For each example, we report temporal relation (first column), generic verbs (second column) and the verbs present in the MECCANO Dataset (third column) which belong to the considered relation.

\subsubsection{Action and Interaction Temporal Annotations}
\label{sec:temporal_ann}
We considered 12 different verbs which describe the actions performed by the participants while building the toy model: \textit{take, put, check, browse, plug, pull, align, screw, unscrew, tighten, loosen} and \textit{fit}. 

We represent each temporal segment as a triplet composed of three different timestamps: 1) the start time which indicates the start of the action, 2) the contact time which indicates the first frame in which the contact between the hand and the object (or between the objects) is clearly visible, changing the state of the objects from \textit{passive} to \textit{active} and 3) the end time of the performed action. We manually annotated both contact and end times of each temporal segment. Since in the MECCANO Multimodal dataset there are three cases in which the action starts before the frame of contact (i.e., take, put and align), for these actions we automatically annotated the start time going back by 0.5 seconds with respect the contact time whereas for the others the start time coincides with the contact.
Only for the \textit{check} verb, where the user doesn't need to touch an object, we annotated the contact time when it is clear from the video sequence that the user is looking at the object (see Figure~\ref{fig:temporal}).
With this procedure, we annotated $8857$ video segments. Since a participant can perform multiple interactions simultaneously, we allowed the annotated segments to overlap (see Figure~\ref{fig:temporal}). In particular, in the MECCANO Multimodal dataset there are 1401 segments (15.82 \%) which overlap with at least another segment. 

We defined $61$ action classes composed of a verb and one or more objects, for example \textit{``align screwdriver to screw''} in which the verb is \textit{align} and the objects are \textit{screwdriver} and \textit{screw}. Depending on the verb and objects involved in the interaction, each temporal segment has been associated to one of the $61$ considered action classes. We analyzed the combinations of our $12$ verb classes and $20$ object classes to find a compact, yet descriptive set of actions classes. The action class selection process has been performed in two stages. In the first stage, we obtained the distributions of the number of active objects generally occurring with each of the $12$ verbs. 
In the second stage, we selected a subset of actions from all combinations of verbs and nouns. 
Let \begin{math} O = \{o_1, o_2, ..., o_n\} \end{math} and \begin{math} V = \{v_1, v_2, ..., v_m\} \end{math} be the set of the object and verb classes respectively.
For each verb $v \in V$, we considered all the object classes $o \in O$ involved in one or more temporal segments labeled with verb $v$. In total, we obtained 61 action classes composing the MECCANO dataset which are shown in Figure~\ref{fig:action_stats}.

\subsubsection{Active Object Bounding Box Annotations}
\begin{figure}[t]
\centering
\includegraphics[width=\columnwidth]{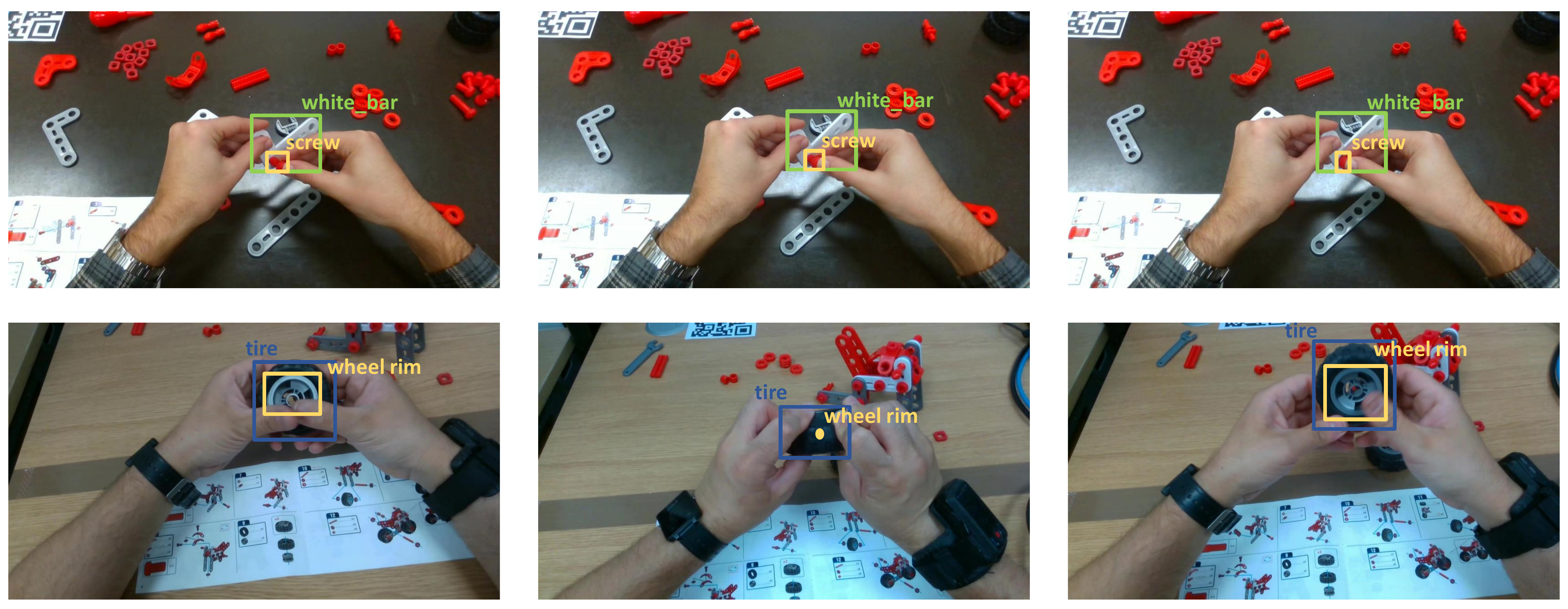}
\caption{Example of bounding box annotations for \textit{active} objects (first row) and occluded \textit{active} objects (second row).}
\label{fig:bbox}
\end{figure}

For each temporal segment, we annotated the \textit{active} objects in frames sampled every $0.2$ seconds. Each active object annotation consists in a \textit{(class, x, y, w, h)} tuple, where \textit{class} represents the class of the object and \textit{(x, y, w, h)} are the 2D coordinates width and height defining the bounding box around the object in the frame. We annotated multiple objects when they were \textit{active} simultaneously (see Figure~\ref{fig:bbox} - first row). If an active object is occluded, even just in a few frames, we annotated it  with a \textit{(class, x, y)} tuple, specifying the class of the object and its estimated 2D position. An example of occluded active object annotation is reported in the second row of Figure~\ref{fig:bbox}. With this procedure, we labeled a total of 64349 frames.

\begin{figure}[t]
\centering
\includegraphics[width=\columnwidth]{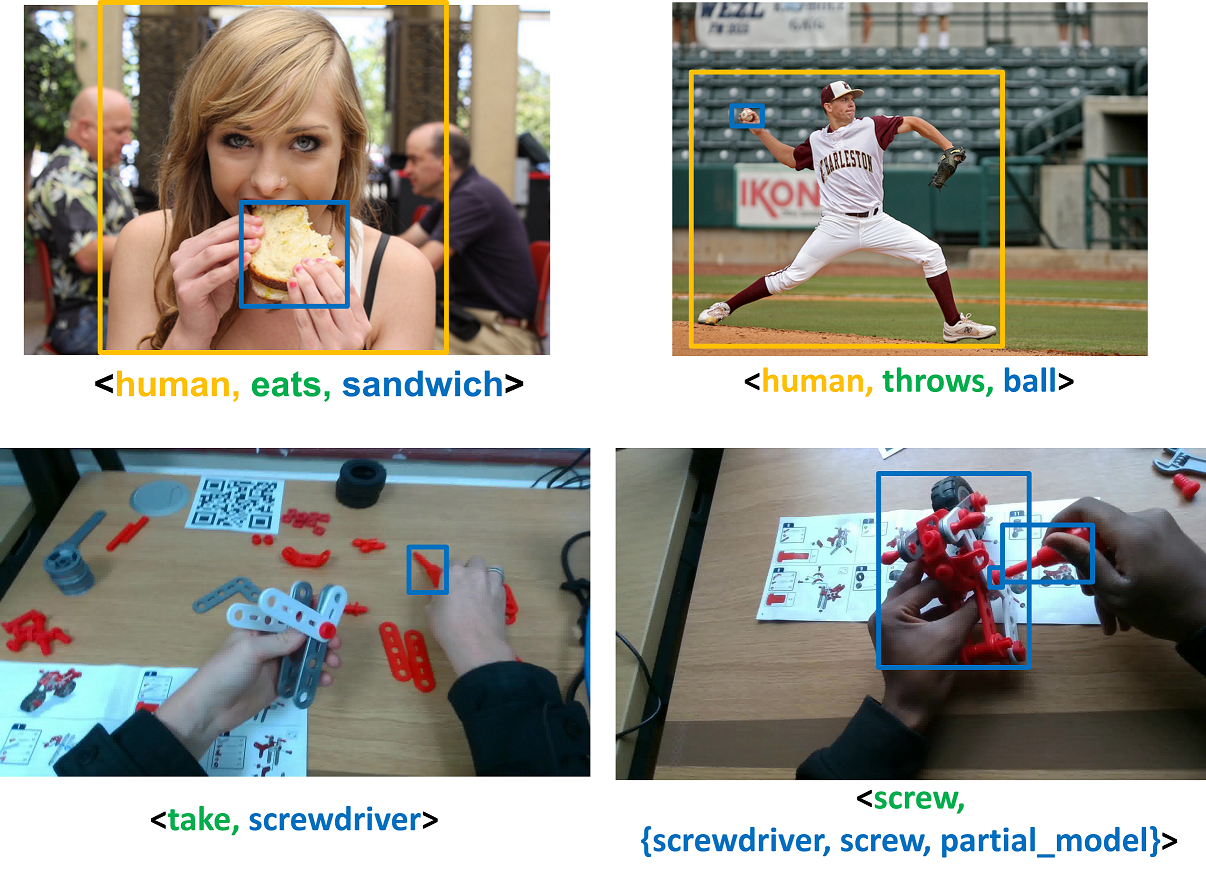}
\caption[]{Examples of Human-Object Interactions from the third point of view (first row) and first point of view (second row)\footnotemark.}
\label{fig:concept_HOI}
\end{figure}

\begin{figure*}[t]
	\centering
	\includegraphics[width=\textwidth]{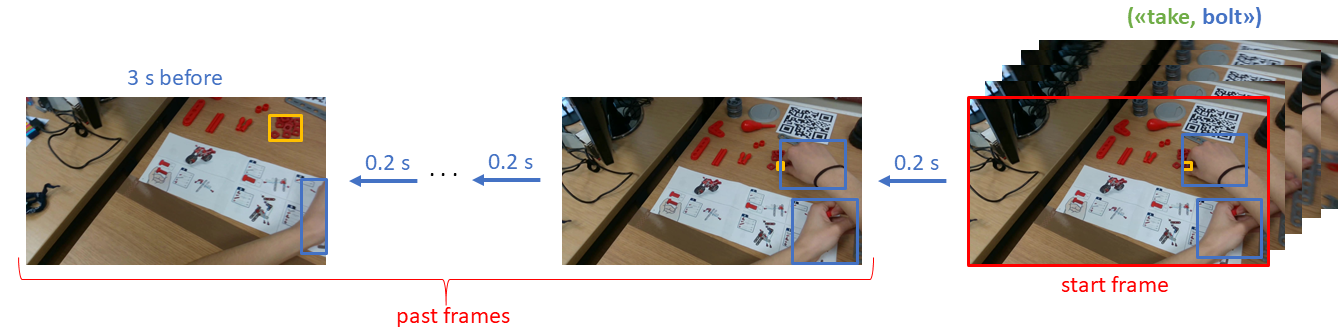}
	\caption{Annotation procedure for next-active objects with bounding boxes in the past frames.}
	\label{fig:next_active_objects}
\end{figure*}

\subsubsection{EHOI Annotations}
\label{sec:EHOI_ann}
The HOI detection task consists in detecting the occurrence of human-object interactions, localizing both the humans taking part in the action and the interacted objects. HOI detection also aims to understand the relationships between humans and objects, which is usually described with a verb.  Possible examples of HOIs are  ``\textit{eat the sandwich}" or ``\textit{throw the ball}" (see Figure~\ref{fig:concept_HOI}-top). 
HOI detection models mostly consider one single object involved in the interaction \cite{Gupta2015VisualSR, HOI_Gupta_09, Gkioxari2018DetectingAR, HOI_Fei_Fei,Chao2018LearningTD}. Hence, an interaction is defined as a triplet in the form \textit{$<$human, verb, object$>$}, where the human is the subject of the action specified by a verb and an object.

Considering the FPV domain, a first formalization of the HOI task has been proposed by \cite{Hands_in_contact_Shan20} who represented an interaction as a triple $<$hand, contact state, object$>$ where, the “contact state” variable assumes one of the following values: (none/self/other/portable/non-portable).
This definition does not describe of the interaction in terms of verb classes and assume that only one object for each hand could be involved in the interaction.
Differently from \cite{Hands_in_contact_Shan20}, we aim to understand more closely human behavior and hence study the Egocentric Human-Object Interaction (EHOI) detection task with the aim of predicting \textit{$<$verb, objects$>$} pairs describing the interaction observed from the egocentric point of view with multiple objects. Note that in EHOIs, the subject is always the camera wearer, so we do not require its localization in the frame, while one or more objects can be involved simultaneously in the interaction. The goal of EHOI detection is to infer the verb and object noun classes, and to localize each active object involved in the interaction. 

Let $O = \{o_1, o_2, ..., o_n\}$ and $V = \{v_1, v_2, ..., v_m\}$ be the sets of objects and verbs respectively. We define an Egocentric Human-Object Interaction $e$ as:

\begin{equation}
\label{eq:1}
e = (v_h, \overline{\rm o_1}, \overline{\rm o_2}, ..., \overline{\rm o_i}\})
\end{equation}

\noindent where \begin{math}v_h \in V\end{math} is the verb characterizing the interaction and \begin{math}{\overline{\rm o_1}, \overline{\rm o_2}, ..., \overline{\rm o_i}} \subseteq O \end{math} are the active objects involved in the interaction. Given the previous definition, we considered all the observed combinations of verbs and objects to represent EHOIs performed by the participants during the acquisition. Two examples are reported in Figure~\ref{fig:concept_HOI})-bottom. Each EHOI annotation is hence composed of a verb annotation and the bounding boxes of \textit{active} objects. Differently from other datasets, MECCANO multimodal has been hence explicitly annotated for the EHOI detection task. 

\subsubsection{Next Active Object Annotations}
\label{sec:nao_annotations}
Due to the limited number of public datasets explicitly annotated to study the future intentions of humans, few past works explored the task of predicting the next-active objects considering the first person point of view \cite{Furnari2017NextactiveobjectPF}, using both RGB and depth signals \cite{Bertasius2017FirstPA}, focusing on the hands \cite{JIANG2021212} or estimating also the time to contact with the future active objects \cite{Ego4D2021}. 

We annotated MECCANO with a set of labels useful to tackle the problem of \textit{Next Active Object} prediction whose goal is to predict and localize the objects that will be involved in a future human-object interaction from the first person view. 

For each human-object interaction, we annotated in the frames preceding the interaction the objects which will be \textit{active objects} in the contact frame. Starting from the contact frame, we sampled frames every 0.2 seconds going backwards up to 3 seconds before the beginning of the temporal segment, or less if there is an overlap with a previous segment\footnote{If an interaction overlaps with previous one, we did not annotate past frames.} (see Figure~\ref{fig:next_active_objects}). Indeed, not all interactions have past frames. For example if the interaction $E_1$ ends at timestamp $T_1$ and the interaction $E_2$ starts at timestamp $T_2 = T_1 + 0.1s$, past frames belonging to the interaction $E_2$ will overlap with frames belonging to the previous interaction $E_1$ and they will not be annotated. With this sampling procedure, we obtained labels in past frames for 75.66\% (6656) of the total number of interactions (8857) present in the dataset. Considering the frames preceding an interaction, each next-active object annotation consists in a \textit{(class, x, y, w, h)} tuple, where \textit{class} represents the class of the object which will be active and the \textit{(x, y, w, h)} tuple defines a bounding box around the considered object. If an object is going to be taken from a pile, then the pile itself is labeled as the next active object. Note that a pile of objects is composed only by objects of the same type. We labeled the pile because we assume that, before a human-object interaction occurs, it is not feasible to infer which object of the pile will be active (see Figure~\ref{fig:next_active_objects}-left).
As in the case of active objects, if an object is occluded, we annotated it with a \textit{(class, x, y)} tuple specifying the class of the object and its estimated 2D position. With this procedure, we labeled a total of 48024 frames with 74127 bounding boxes\footnote{See supplementary material for additional details.}. 

\begin{table*}[]
	\caption{Statistics of the three splits: Train, Validation and Test. The 4th column indicates the percentage of videos belonging to the related subset with respect to the total number of videos present in the MECCANO Dataset.}
	\label{tab:splits}
	\centering
	\resizebox{\textwidth}{!}{%
		\begin{tabular}{l|ccccccccc}
			\multicolumn{1}{c|}{\textbf{Split}} & \textbf{\#Videos} & \textbf{Duration (min)} & \textbf{\%} & \textbf{\#EHOIs Segments} & \textbf{Obj. BBoxes} & \textbf{Hands BBoxes} & \textbf{NAO BBoxes} & \textbf{Country (U.K/Italy)} & \textbf{Table (Light/Dark)} \\ \hline
			Train & 11 & 236.47 & 55\% & 5057 & 37386 & 96556 & 28152 & 6/5 & 6/5 \\ 
			Val & 2 & 46.57 & 10\% & 977 & 6983 & 19636 & 5490 & 1/1 & 1/1 \\ 
			Test & 7 & 134.93 & 35\% & 2824 & 19980 & 87924 & 14382 & 4/3 & 4/3 \\ \hline
		\end{tabular}%
	}
\end{table*}

\subsubsection{Hands Annotations}
As MECCANO multimodal features actions and human-object interactions from the first person point of view where the hand of the user are visible during the objects manipulation, knowledge on the position of the hands could be an important modality to explore.
For each temporal segment, we annotated the hands of the participants with a bounding box on the set of frames belonging to the interaction (i.e. from the start frame to the end frame) and in the past frames preceding the interaction. Each hand annotation consists in a \textit{(class, x, y, w, h)} tuple, where \textit{class} represents the side of the hand (i.e. left or right) and \textit{(x, y, w, h)} defines a bounding box around the considered hand. We split this labeling procedure in two stages. Firstly, we processed the frames with the Hand Object Detector introduced in \cite{Hands_in_contact_Shan20}. This detector infers if an hand is involved in an interaction through the contact with active objects. In particular, the detector predicts the hand location, the side, a contact state, and a box around the object in contact. We considered only the hand location and the side for each of the processed frame. In the second stage, annotators checked if the predicted bounding boxes and the associated class were correct and adjusted them or added a new annotation if there was a missing hand prediction. If the bounding box was not precise or the class was wrong, they refined the bounding box and corrected the class of the hand. With this procedure, we annotated \textit{89628} frames with \textit{169625} bounding boxes around the hands. See supplementary material for additional details.
We used this set of annotations as additional modality to tackle the \textit{Action Anticipation} task as described in Section~\ref{sec:Action_anticipation}. Hands annotations could be useful also to understand human-object interactions or to recognize the actions performed by the user.

\section{Benchmarks and Baseline Results}
\label{sec:benchmark}

\begin{figure*}[t]
	\centering
	\includegraphics[width=\textwidth]{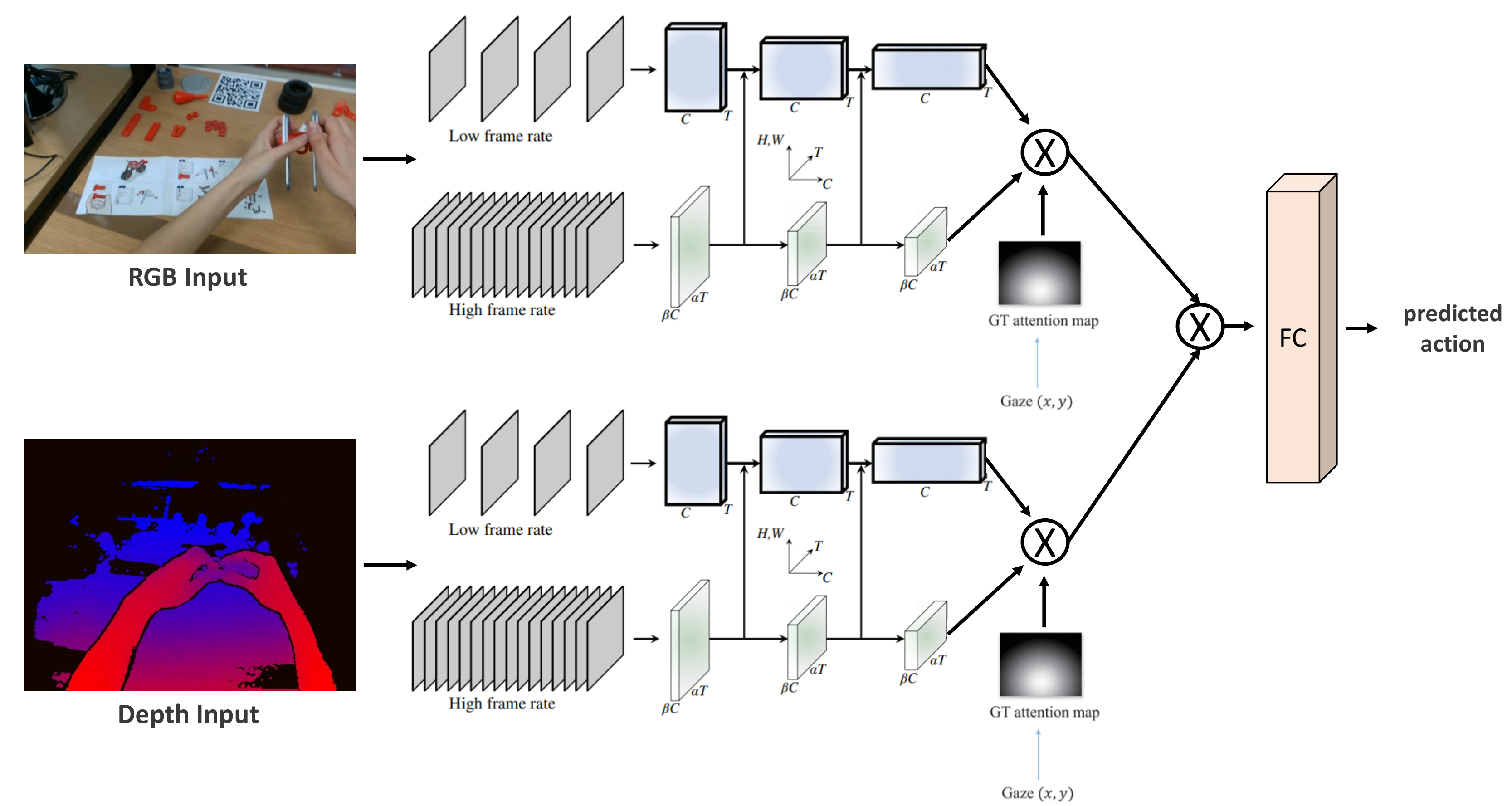}
	\caption{SlowFast (RGB-Depth-Gaze) architecture which gets as input three different signals (RGB, depth and gaze).}
	\label{fig:action_rec_architecture}
\end{figure*}

The MECCANO dataset is suitable to study a variety of tasks, considering its multimodality and the challenging industrial-like scenario in which it was acquired. In this paper, we proposed five tasks related to human's behavior understanding and provide baseline results: 1) \textit{Action Recognition}, 2) \textit{Active Object Detection and Recognition}, 3) \textit{Egocentric Human-Object Interaction (EHOI) Detection}, 4) \textit{Action Anticipation} and 5) \textit{Next Active Object Detection}. While some of these tasks have been studied in previous works, none of them has been studied in industrial scenarios from the egocentric perspective also considering multimodal observations. Moreover, there are only few datasets publicy available \cite{Damen2020RESCALING, Ego4D2021, Sener2022Assembly101AL} which can be used to study different tasks simultaneously and to develop a complete system for human behavior understanding taking into account different aspects (e.g., actions, interactions, objects, future intentions).

MECCANO has been split into three subsets (\textit{Training, Validation} and \textit{Test}) designed to balance the different types of desks (light, dark) and countries in which the videos have been acquired (IT, U.K.). We used the training set to train the baselines, the validation set for hyperparameters tuning and the test set to test the trained models. Each video has been entirely assigned to one of the three different subsets.
Table~\ref{tab:splits} reports some statistics about the three splits, such as the number of videos, the total duration (in seconds), the number of temporally annotated EHOIs and the number of bounding box annotations.

\begin{figure*}[t]
	\centering
	\includegraphics[width=\textwidth]{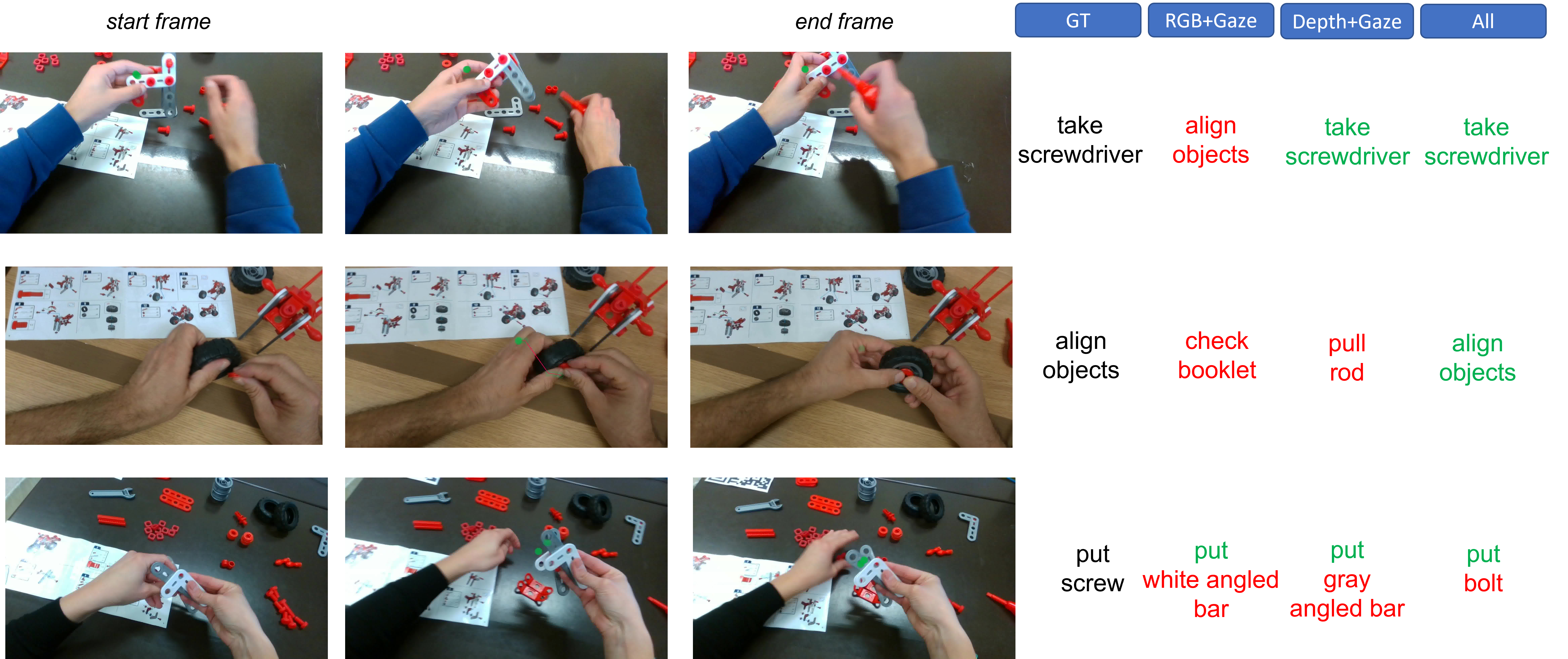}
	\caption{Qualitative results of the action recognition models. Green and red colors represent the correct/wrong predictions for both verbs and objects which compose the action class. In the first column, we report the ground truth action class, in the second one the prediction of the model SlowFast (RGB-Gaze), in third column the action class predicted by SlowFast (Depth-Gaze) and in the last column (\textit{All}) the prediction of SlowFast (RGB-Depth-Gaze).}
	\label{fig:action_rec_qual_res}
\end{figure*}

\begin{table*}[]
	\caption{Results for the action recognition task. The best results are reported in bold, whereas the second best results are underlined.}
	\label{tab:action_rec}
	\resizebox{\textwidth}{!}{%
		\begin{tabular}{l|ccccc}
			\textbf{Method} & \multicolumn{1}{l}{\textbf{Top-1 Accuracy}} & \multicolumn{1}{l}{\textbf{Top-5 Accuracy}} & \multicolumn{1}{l}{\textbf{AVG Class Precision}} & \multicolumn{1}{l}{\textbf{AVG Class Recall}} & \multicolumn{1}{l}{\textbf{AVG F1-score}} \\ \hline
			SlowFast (RGB) & 45.16 & 73.75 & 50.33 & 45.16 & 46.66 \\
			SlowFast (Depth) & 45.13 & 72.19 & 50.28 & 45.13 & 46.96 \\
			SlowFast (RGB-Depth) & \underline{49.49} & \underline{77.61} & \underline{56.13} & \underline{49.49} & \underline{51.90} \\
			SlowFast (RGB-Gaze) & 45.34 & 73.61 & 49.83 & 44.60 & 46.25 \\
			SlowFast (Depth-Gaze) & 45.27 & 72.30 & 50.62 & 45.27 & 47.23 \\
			SlowFast (RGB-Depth-Gaze) & \textbf{49.66} & \textbf{77.82} & \textbf{56.69} & \textbf{49.66} & \textbf{52.25}
		\end{tabular}
	}
\end{table*}

\subsection{Action Recognition}
\label{sec:action_recognition}
Action Recognition consists in determining the action performed by the camera wearer from the observation of an egocentric video segment. Specifically, let \begin{math}C_a = \{c_1, c_2, ..., c_n\} \end{math} be the set of action classes and let \begin{math}A_i = [t_{s_i}, t_{e_i}]\end{math} be the video segment,  where $t_{s_i}$ and $t_{e_i}$ are the start and the end times of the action respectively. The aim is to assign the correct action class $c_i \in C_a$ to the segment $A_i$. 
We evaluate action recognition using Top-1 and Top-5 accuracy computed on the whole test set. As class-aware measures, we report class-mean precision, class-mean recall and $F_1$-score.

\subsubsection{Baseline}
\label{sec:baseline_act_rec}
As a baseline we considered SlowFast \cite{feichtenhofer2018slowfast}, which is a state-of-the-art method for action recognition. To explore the usefulness of multimodal signals for this task, we adopted different instances of the SlowFast architecture, as detailed in the following.\\
\textbf{SlowFast (RGB):\hspace{1mm}} the 3D network architecture which takes as input the RGB clip to perform action recognition, as implemented in Pyslowfast \cite{fan2020pyslowfast}.\\
\textbf{SlowFast (Depth):\hspace{1mm}} a 3D network architecture which takes as input the clip composed of depth maps related to the corresponding RGB frames.\\
\textbf{SlowFast (RGB-Depth):\hspace{1mm}} this model is composed of two SlowFast networks which process different input signals (RGB and the corresponding Depth frames). The two probability distributions obtained as output are averaged to obtain the final action prediction. \\ 
\textbf{SlowFast (RGB-Gaze):\hspace{1mm}} inspired by \cite{deep_attention_Minlong}, we integrated human gaze into the SlowFast network. Specifically, we used the ground truth gaze fixation to obtain an attention map which focuses on the most relevant spatial regions of the video frames along the time dimension. This attention map is multiplied with the output feature maps of both the slow and fast pathways. Then, we fused the new combined feature maps by concatenation and fed to the fully connected layers to obtain the final prediction.\\
\textbf{SlowFast (Depth-Gaze):\hspace{1mm}} this model is similar to SlowFast (RGB-Gaze) but it takes as input the depth maps of the video clips rather than RGB frames.\\
\textbf{SlowFast (RGB-Depth-Gaze):\hspace{1mm}} this model is composed of two instances of SlowFast model (one for each input signal) and integrates human gaze as seen in SlowFast (RGB-Gaze) architecture (Figure~\ref{fig:action_rec_architecture}).

\subsubsection{Results}
Table~\ref{tab:action_rec} reports the results obtained with the adopted baselines for the action recognition task. Baselines using only one modality (RGB or Depth), achieve similar performance in terms of Top-1 Accuracy (45.16\% vs. 45.13\%) and AVG F1-score (46.66\% vs. 46.96\%). Fusing the RGB and Depth signals, we obtain better results with respect to all the baselines which use one or two modalities. Exploiting all the signals present in the MECCANO Dataset (RGB, Depth and Gaze), we obtain the best results considering all the evaluation measures (last row of Table~\ref{tab:action_rec}).
Even if the gaze modality represents an additional signal to guide learning, improvements with this modality are minor. The limited improvement obtained using this modality with respect to the model which uses RGB and Depth signals, could be related to the nature of the adopted architecture, which is simple and can be optimized in future works. Qualitative results are reported in Figure~\ref{fig:action_rec_qual_res}, which highlights how the use of multiple signals improve the performance on the action recognition tasks (first and second row). 
In general, the results suggest that the use of multiple modalities allows to improve the results in the MECCANO dataset. Moreover, the dataset is a challenging testbed for action recognition and offers a new scenario to compare Classic and Multimodal action recognition algorithms.

\subsection{Active Object Detection and Recognition}
\label{sec:AODR}
Differently from previous works, we consider two distinct but related tasks: active object detection and active object recognition.
In some cases, detecting the active objects manipulated by the human without considering the object class \cite{Hands_in_contact_Shan20} can be useful, for example when the taxonomy of the object classes is difficult to obtain or to initialize a tracker \cite{TREK150}. However, when a taxonomy is available, predicting the object classes can enable practical applications (e.g., monitoring the usage time of specific objects).

The aim of the Active Object Detection task is to detect all the \textit{active} objects. 
Let \begin{math} O_{act} = \{o_1, o_2, ..., o_n\} \end{math} be the set of \textit{active} objects in a given frame. The goal is to detect with a bounding box each \textit{active} object $o_i \in O_{act}$. As evaluation measure, we use the Average Precision~(AP), which is used in standard object detection benchmarks. We set the IoU threshold equal to~$0.5$ in our experiments as in the standard Pascal VOC mAP measure \cite{PascalVOC_Zisserman_15}.

The active object recognition task instead, consists in detecting and recognizing the \textit{active} objects involved in EHOIs considering the $20$ object classes of the MECCANO dataset.
Formally, let \begin{math} O_{act} = \{o_1, o_2, ..., o_n\}\end{math} be the set of \textit{active} objects in the image and let \begin{math} C_{o} = \{c_1, c_2, ..., c_m\} \end{math} be the set of object classes. The task consists in detecting objects $o_i \in O_{act}$ and assigning them the correct class label $c_i \in C_{o}$.
We use mAP \cite{PascalVOC_Zisserman_15} with an IoU threshold equal to $0.5$ for the evaluations.

\begin{table}[]
	\caption{Baseline results for the \textit{active} object detection task.}
	\label{tab:active_det}
	\centering
	\resizebox{\columnwidth}{!}{%
		\begin{tabular}{l|c}
			\multicolumn{1}{c|}{\textbf{Method}} & \multicolumn{1}{l}{\textbf{AP (IoU \textgreater 0.5)}} \\ \hline
			Hand Object Detector \cite{Hands_in_contact_Shan20} & 11.17\% \\
			Hand Object Detector \cite{Hands_in_contact_Shan20} (Avg dist.)  & 11.10\% \\
			Hand Object Detector \cite{Hands_in_contact_Shan20} (All dist) & 11.34\% \\
			Hand Object Detector \cite{Hands_in_contact_Shan20} + Objs re-training  & 20.18\% \\
			Hand Object Detector \cite{Hands_in_contact_Shan20} + Objs re-training (Avg dist.)  & 33.33\% \\
			Hand Object Detector \cite{Hands_in_contact_Shan20} + Objs re-training (All dist.)  & \textbf{38.14\%} \\ \hline
		\end{tabular}%
	}
\end{table}

\subsubsection{Methods}
To address the problem of detecting active objects, the Hand-Object Detector proposed in \cite{Hands_in_contact_Shan20} has been considered as a baseline. The model has been designed to detect hands and objects when they are in contact with hands. This architecture is based on Faster-RCNN \cite{ren2015faster} and predicts a box around the visible human hands, as well as boxes around the objects the hands are in contact with and a link between them. We used the Hand-Object Detector \cite{Hands_in_contact_Shan20} pretrained on EPIC-Kitchens \cite{Damen2018EPICKITCHENS}, EGTEA \cite{Li2018_EGTEA-GAZE+} and CharadesEGO \cite{Sigurdsson2018Charades} as provided by the authors \cite{Hands_in_contact_Shan20}. For our purpose the model has been trained to recognize hands and to detect the \textit{active} objects regardless of their class.
With default parameters, the Hand-Object Detector can find at most two \textit{active} objects in contact with hands. Since our dataset tends to contain more \textit{active} objects in a single EHOI (up to 7), we consider two variants of this model by changing the threshold on the distance between hands and detected objects. In the first variant, the threshold is set to the average distance between hands and \textit{active} objects in the MECCANO dataset. We named this variant ``\textit{Avg distance}''. In the second variant, we removed the thresholding operation and considered all detected objects as \textit{active} objects. We named this variant ``\textit{All objects}''. 
We further adapted the Hand-Object Detector \cite{Hands_in_contact_Shan20} re-training the Faster-RCNN component to detect all \textit{active} objects of the MECCANO dataset. Faster-RCNN has been trained on the training and validation sets using the provided \textit{active} object class labels. Since the task aims to only localize objects, we discard predicted object classes at test time.

For the active object recognition task, as a baseline, we used a standard Faster-RCNN \cite{ren2015faster} object detector. For each image, the object detector predicts \textit{(x, y, w, h, class)} tuples which represent the object bounding boxes and the associated classes. We used the same Faster-RCNN model adopted for the Active Object Detection task, retaining also object classes at test time.

\begin{table}[]
\caption{Baseline results for the \textit{active} object recognition task.}
\label{tab:active_rec}
\centering
\resizebox{0.6\columnwidth}{!}{%
\begin{tabular}{clc}
\multicolumn{1}{l|}{\textbf{ID}} & \multicolumn{1}{c|}{\textbf{Class}} & \textbf{AP (per class)} \\ \hline
\multicolumn{1}{c|}{0} & \multicolumn{1}{l|}{instruction booklet} & 46.18\% \\
\multicolumn{1}{c|}{1} & \multicolumn{1}{l|}{gray\_angled\_perforated\_bar} & 09.79\% \\
\multicolumn{1}{c|}{2} & \multicolumn{1}{l|}{partial\_model} & 36.40\% \\
\multicolumn{1}{c|}{3} & \multicolumn{1}{l|}{white\_angled\_perforated\_bar} & 30.48\% \\
\multicolumn{1}{c|}{4} & \multicolumn{1}{l|}{wrench} & 10.77\% \\
\multicolumn{1}{c|}{5} & \multicolumn{1}{l|}{screwdriver} & 60.50\% \\
\multicolumn{1}{c|}{6} & \multicolumn{1}{l|}{gray\_perforated\_bar} & 30.83\% \\
\multicolumn{1}{c|}{7} & \multicolumn{1}{l|}{wheels\_axle} & 10.86\% \\
\multicolumn{1}{c|}{8} & \multicolumn{1}{l|}{red\_angled\_perforated\_bar} & 07.57\% \\
\multicolumn{1}{c|}{9} & \multicolumn{1}{l|}{red\_perforated\_bar} & 22.74\% \\
\multicolumn{1}{c|}{10} & \multicolumn{1}{l|}{rod} & 15.98\% \\
\multicolumn{1}{c|}{11} & \multicolumn{1}{l|}{handlebar} & 32.67\% \\
\multicolumn{1}{c|}{12} & \multicolumn{1}{l|}{screw} & 38.96\% \\
\multicolumn{1}{c|}{13} & \multicolumn{1}{l|}{tire} & 58.91\% \\
\multicolumn{1}{c|}{14} & \multicolumn{1}{l|}{rim} & 50.35\% \\
\multicolumn{1}{c|}{15} & \multicolumn{1}{l|}{washer} & 30.92\% \\
\multicolumn{1}{c|}{16} & \multicolumn{1}{l|}{red\_perforated\_junction\_bar} & 19.80\% \\
\multicolumn{1}{c|}{17} & \multicolumn{1}{l|}{red\_4\_perforated\_junction\_bar} & 40.82\% \\
\multicolumn{1}{c|}{18} & \multicolumn{1}{l|}{bolt} & 23.44\% \\
\multicolumn{1}{c|}{19} & \multicolumn{1}{l|}{roller} & 16.02\% \\ \hline
\multicolumn{1}{l}{} &  & \multicolumn{1}{l}{} \\ \cline{2-3} 
\multicolumn{1}{l}{} & \multicolumn{1}{c|}{\textbf{mAP}} & 30.39\%
\end{tabular}%
}
\end{table}

\begin{figure}[t]
	\centering
	\includegraphics[width=\columnwidth]{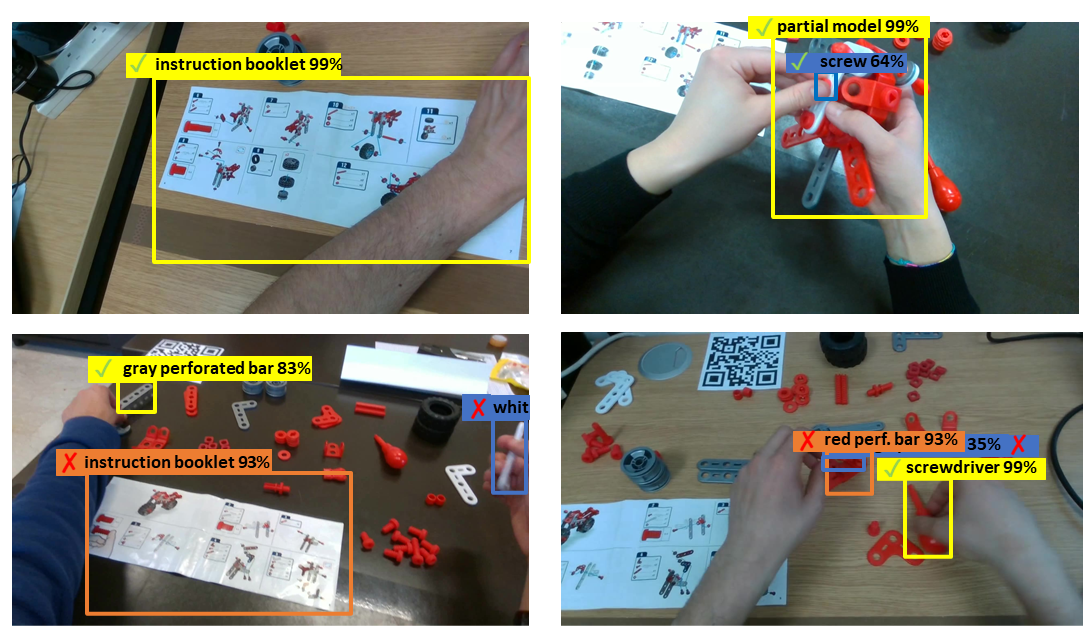}
	\caption{Qualitative results for the Active Object Recognition task.}
	\label{fig:active_objects_qual}
\end{figure}

\subsubsection{Results}

\begin{figure*}[t]
	\centering
	\includegraphics[width=\textwidth]{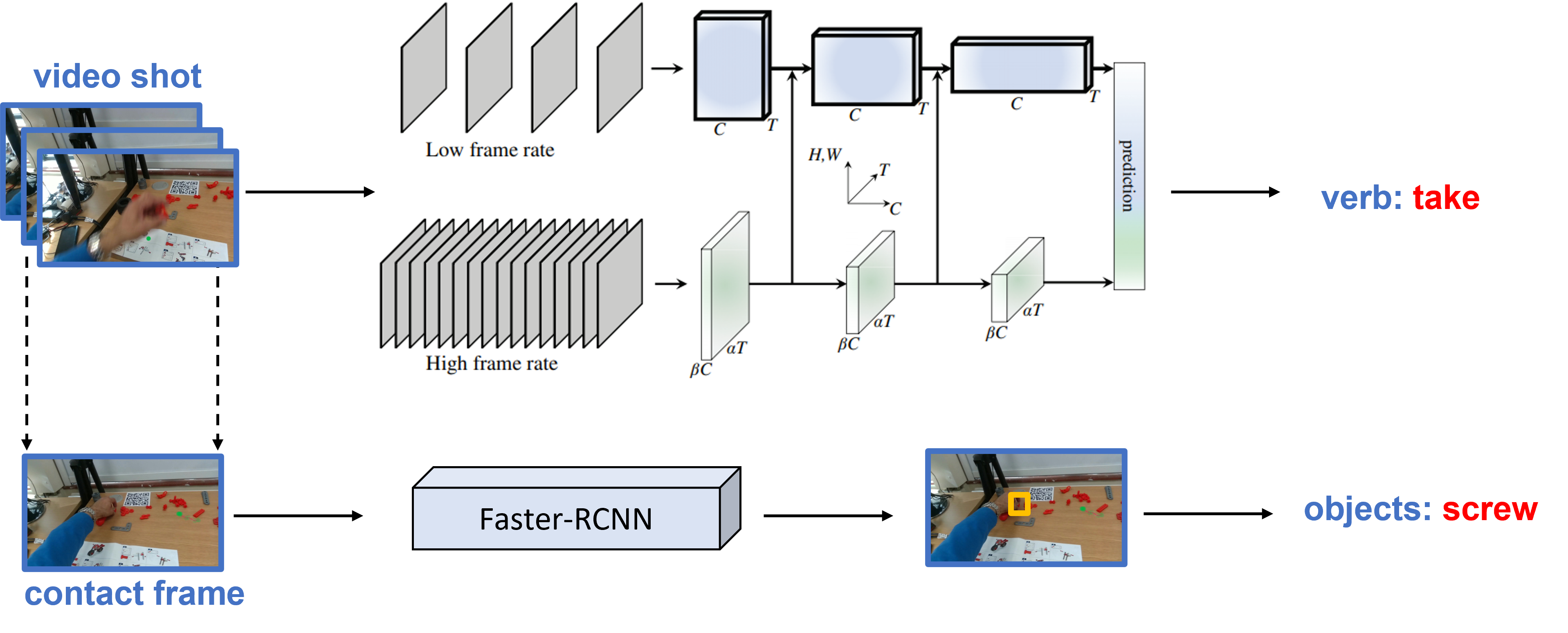}
	\caption{Proposed architecture for the EHOI detection task.}
	\label{fig:EHOI_arch}
\end{figure*}

\begin{table*}[]
\caption{Results on the EHOI Detection task. The best results are reported in bold, whereas the second best results are underlined.}
\label{tab:EHOI}
\resizebox{\textwidth}{!}{
\begin{tabular}{lccc|ccc}
 & \multicolumn{3}{c|}{\textbf{mAP\textsubscript{verb}}} & \multicolumn{3}{c}{\textbf{mAP\textsubscript{verb,noun}}} \\
\textbf{Method} & \textbf{IoU@50} & \textbf{IoU@30} & \textbf{IoU@10} & \textbf{IoU@50} & \textbf{IoU@30} & \textbf{IoU@10} \\ \hline
SlowFast (RGB) + Faster-RCNN & 26.44 & 28.83 & 30.45 & 19.14 & 21.36 & 22.05 \\
SlowFast (Depth) + Faster-RCNN & \textbf{29.10} & \textbf{31.81} & \textbf{33.81} & \underline{21.37} & \underline{24.01} & \underline{25.01} \\
SlowFast (RGB-Depth) + Faster-RCNN & 26.49 & 28.88 & 30.51 & 19.20 & 21.42 & 22.12 \\
SlowFast (RGB-Gaze) + Faster-RCNN & 28.82 & \underline{ 31.56} & \underline{33.55} & \textbf{21.38} & \textbf{24.04} & \textbf{25.04} \\
SlowFast (Depth-Gaze) + Faster-RCNN & \underline{28.92} & 31.51 & 33.38 & 20.79 & 23.28 & 24.20 \\
SlowFast (RGB-Depth-Gaze) + Faster-RCNN & 28.51 & 31.10 & 32.97 & 20.65 & 23.12 & 24.03
\end{tabular}
}

\end{table*}

\begin{figure}[]
	\centering
	\includegraphics[width=\columnwidth]{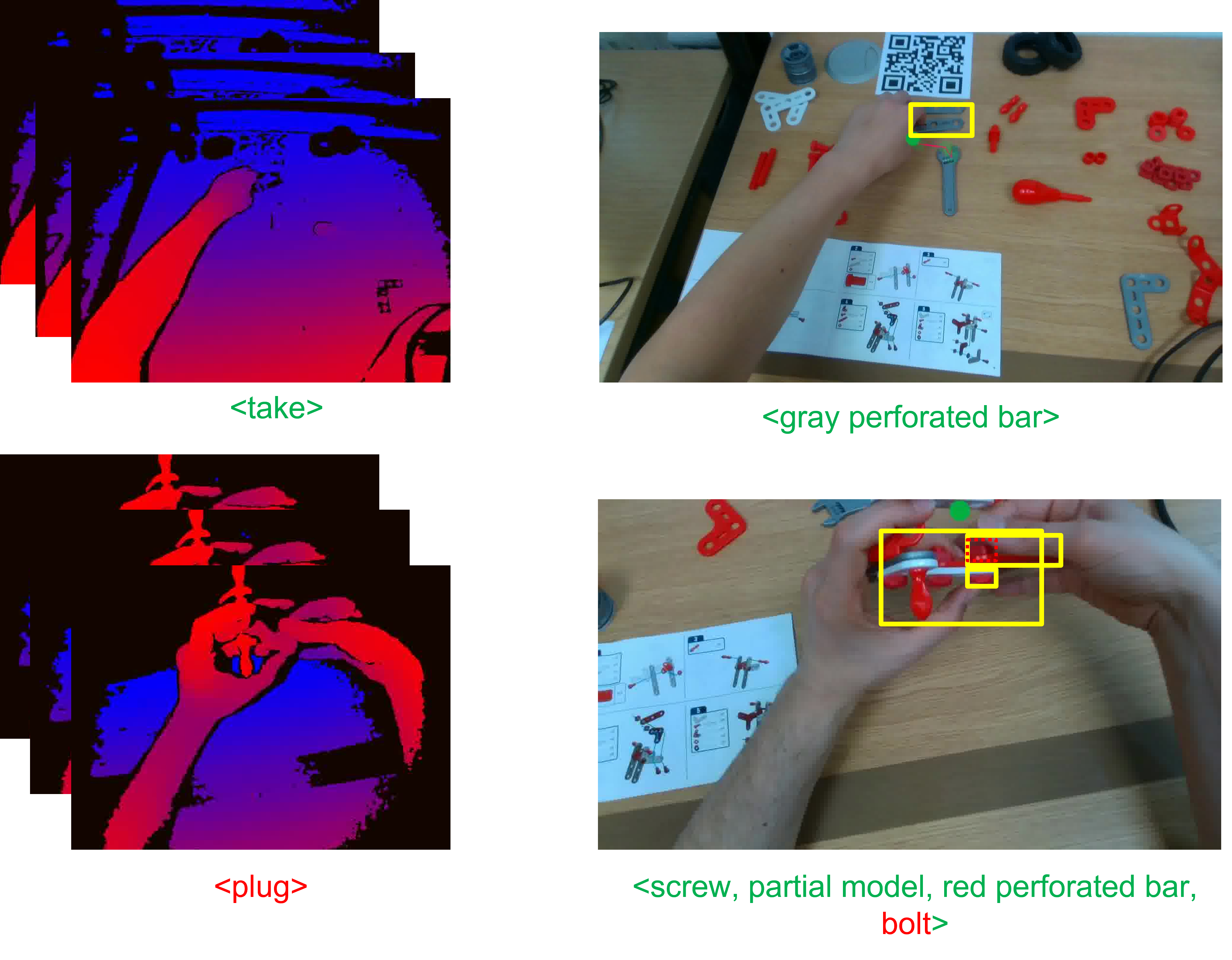}
	\caption{Qualitative results of the SlowFast (Depth) + Faster-RCNN method for the EHOI Detection task. On the left, the SlowFast(Depth) verb prediction, while, on the right, the active objects detected by Faster-RCNN. We reported wrong verb predictions and missing object detection with a dashed red bounding box.}
	\label{fig:ehoi_qual}
\end{figure} 

Table~\ref{tab:active_det} shows the results obtained by the \textit{active} object detection task baselines. The results highlight that the Hand-Object Detector \cite{Hands_in_contact_Shan20} is not able to generalize to the challenging domain offered by MECCANO Multimodal. All the three variants of the Hand-Object Detector using the original object detector obtained an AP approximately equal to 11\% (first three rows of Table~\ref{tab:active_det}). Re-training the object detector on the MECCANO dataset allowed to improve performance by significant margins. In particular, using the standard distance threshold value, we obtained an AP of 20.18\%. If we consider the average distance as the threshold to discriminate \textit{active} and \textit{passive} objects, we obtain an AP of 33.33\%. Removing the distance threshold (last row of Table~\ref{tab:active_det}), allows to outperform all the previous results obtaining an AP equal to 38.14\%. Note that, since no distance threshold is considered, the baseline consists in just using a Faster R-CNN object detector trained on the target context. This suggests that adapting the general object detector to the challenging domain of the proposed dataset is key to performance. Indeed, training the object detector to detect only \textit{active} objects in the scene already allows to obtain reasonable results, while there is still space for improvement.
Table~\ref{tab:active_rec} reports the AP values obtained by the Faster R-CNN active object detection baseline for each class considering all the videos belonging to the test set of MECCANO. The last column shows the average of the AP values for each class and the last row reports the mAP value for the test set. The mAP was computed as the average of the mAP values obtained in each test video. AP values in the last column show that large objects are easier to recognize (e.g. \textit{instruction booklet: 46.48\%; screwdriver: 60.50\%; tire: 58.91\%; rim: 50.35\%}). Results suggests that the proposed dataset is challenging due to the presence of small objects. We reported qualitative results in Figure~\ref{fig:active_objects_qual}.
We leave the investigation of more specific approaches to active object detection to future studies.

\begin{figure*}[t]
	\centering
	\includegraphics[width=\textwidth]{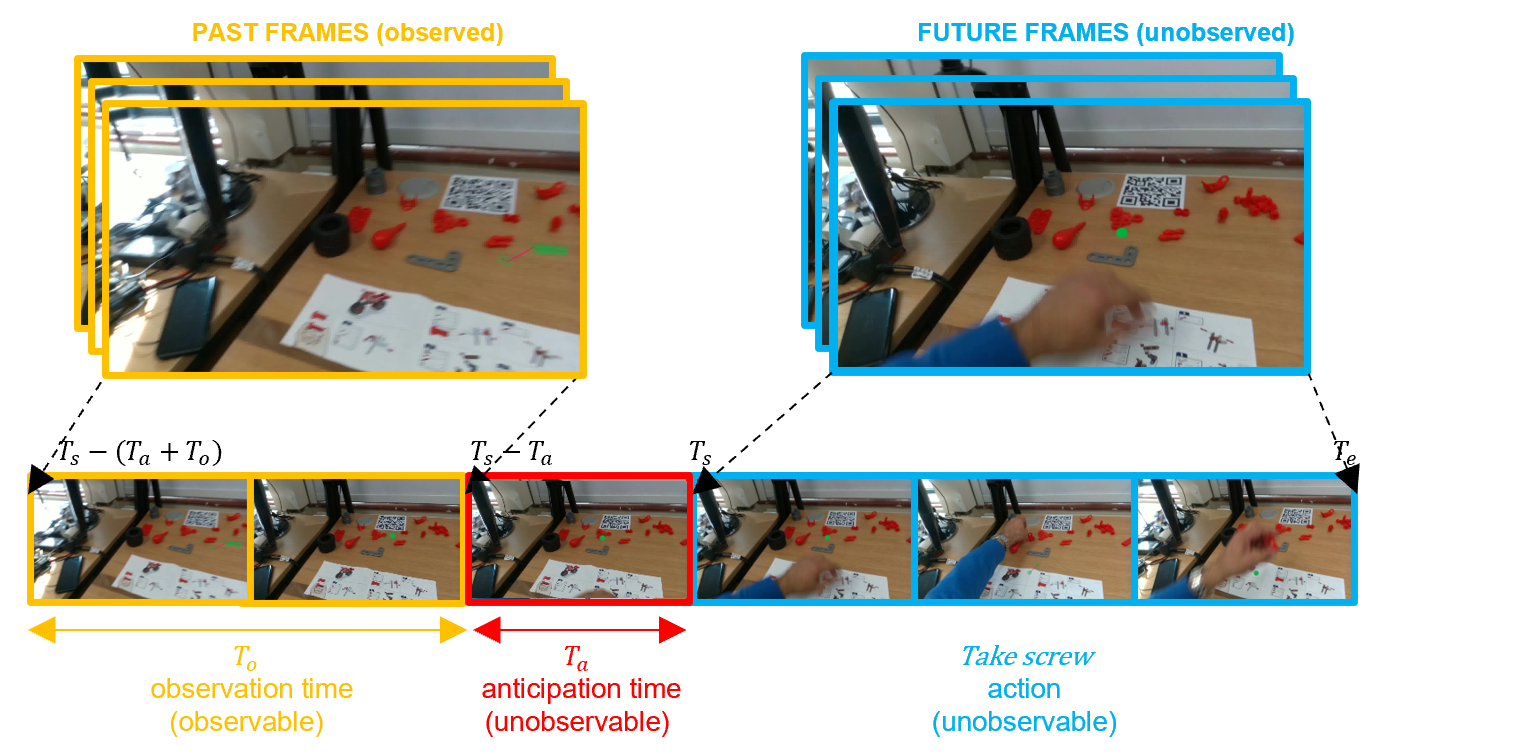}
	\caption{The goal of the action anticipation task is to predict egocentric actions from an observation of the past.}
	\label{fig:concept_action_ant}
\end{figure*}

\subsection{EHOI Detection}
\label{sec:EHOI}
The goal of this task is to detect all egocentric human-object interactions (EHOI) in contact frames. As denoted in the definition of EHOIs as $<$verb, objects$>$ pairs (see Equation~\ref{eq:1}), methods should  detect and recognize all the \textit{active} objects in the scene, as well as the verb describing the action performed by the human.
Following previous works \cite{Gupta2015VisualSR, Gkioxari2018DetectingAR}, \textit{``AP\textsubscript{role}''} is used as evaluation measure for this task.
Formally, a detected EHOI is considered as a true positive if 1) the predicted object bounding box has a IoU of 0.5 or higher with respect to a ground truth annotation and 2) the predicted verb matches with the ground truth. 
Note that only the \textit{active} object bounding box location (not the correct class) is considered in this measure.
Since we want to also recognize \textit{active} objects, we consider a variant of \textit{``AP\textsubscript{role}''} adding the following condition: 3) the predicted object class matches with the ground truth. We called this measure \textit{``AP\textsubscript{verb,noun}''}\footnote{Note that \textit{``AP\textsubscript{noun}''} which considers only the 1) and 3) conditions is the same measure computed in Table~\ref{tab:active_rec}.}. We used both measures to evaluate our method. To better highlight the difference between the two measures we will refer to the \textit{``AP\textsubscript{role}''} as \textit{``AP\textsubscript{verb}''}. 
Moreover, we used different values of IoU (i.e., 0.5, 0.3 and 0.1) to compute the different \textit{``AP''} values.

\subsubsection{Method}
Our baseline is based on the combination of a SlowFast network \cite{fan2020pyslowfast} trained to predict the verb of the EHOI considering a video clip sampled around the frame following the interaction temporal annotations, and Faster-RCNN \cite{ren2015faster}, which detects and recognizes all \textit{active} objects in the frame as shown in Figure~\ref{fig:EHOI_arch}.
Similar to the action recognition task, we explored the potential of the multimodal signals present in the MECCANO dataset, considering different instances of the SlowFast network which relies to RGB, Depth and Gaze signals similarly to what described in Section~\ref{sec:action_recognition}. Note that, in this case, the SlowFast network has been trained on the 12 verb classes of the MECCANO dataset which describe the interactions performed by the users rather than the action classes as done in Section~\ref{sec:action_recognition}. Moreover, due to the difference related to the \textit{action} and \textit{interaction} concepts, we used the EHOI annotations which are slightly different in terms of temporal boundaries with respect to the action annotations as detailed in Section~\ref{sec:EHOI_ann}. For the object detector component, we used the same model trained for the active object recognition task.

\begin{figure*}[t]
	\centering
	\includegraphics[width=\textwidth]{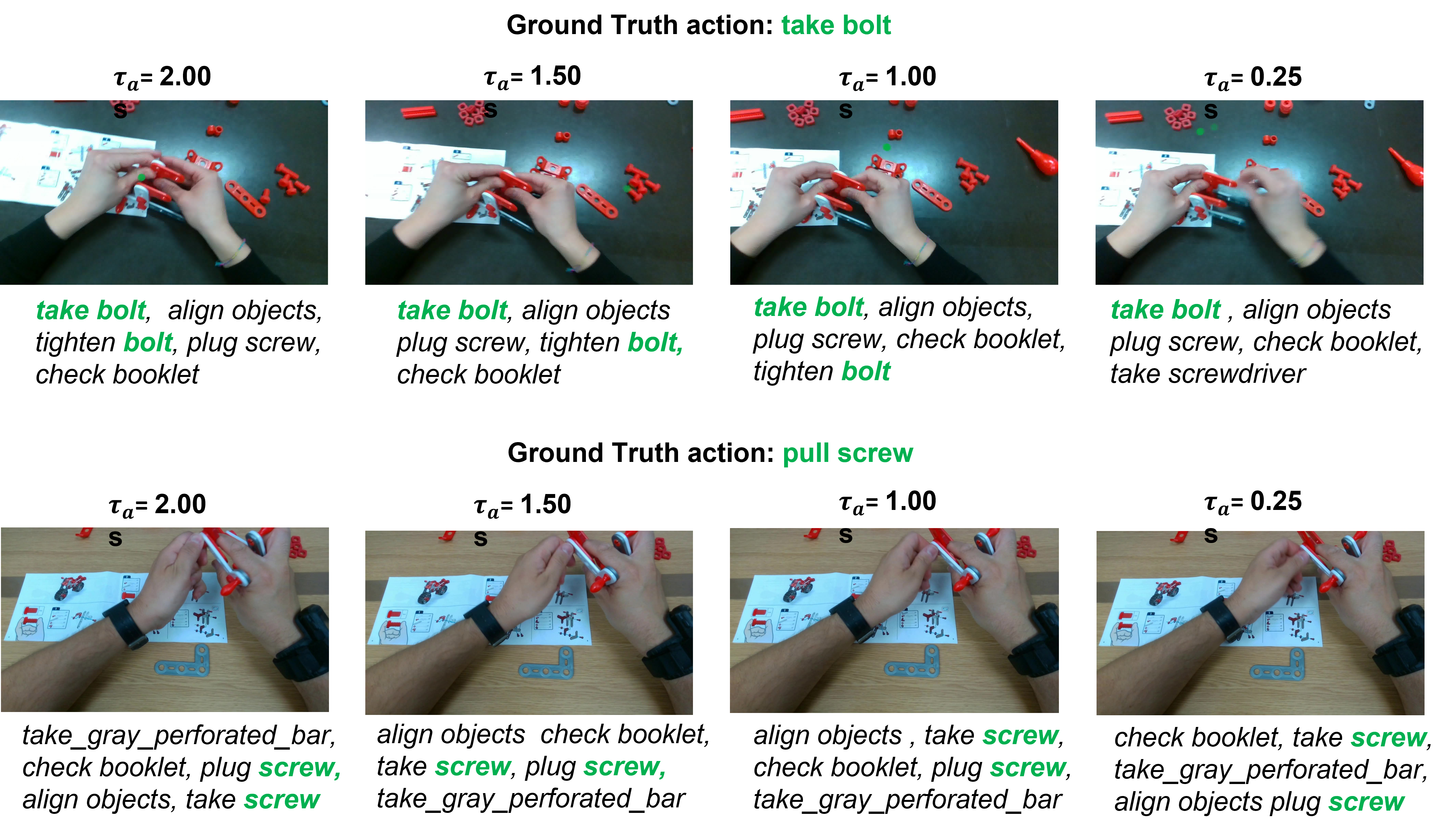}
	\caption{Qualitative results of the proposed approach based on RULSTM which is composed of three branches: gaze, objects and hands.}
	\label{fig:act_ant_qual}
\end{figure*}

\subsubsection{Results}
Table~\ref{tab:EHOI} reports the results for the EHOI detection task. The baseline which considers only the depth signal (second row) obtained the best results for all the mAP\textsubscript{verb} measures with different values of IoU (29.10, 31.81 and 33.81). This can be motivated by the fact that the depth signal (second row) helps to focus the attention on the hands which are most relevant to understand the motion and discriminate the different verbs. Moreover, the 3D network does not aim to predict the object classes, which makes the RGB signal less useful. Interestingly, the use of the depth signal and the attention map computed from the gaze signal (fifth row) does not help for the recognition of the interaction (29.10 versus 28.92) while using the attention map of the gaze with the RGB signal (fourth row) improves the recognition of the interactions obtaining the second best performance considering the IoU@30 and IoU@10 measures, with maP values of 31.56 and 33.55 respectively. 
If we consider the mAP\textsubscript{verb,noun} in which also the object class needs to be correctly predicted, the baseline which considers RGB and Gaze signals (4th row) obtained best results considering all the IoU values (21.38, 24.04 and 25.04). In general, the mAP\textsubscript{verb,noun} values are lower than  mAP\textsubscript{verb} values because we also consider the condition of predicting the correct object class. Qualitative results of SlowFast (Depth) + Faster-RCNN are reported in Figure~\ref{fig:ehoi_qual}.
Despite the promising performance of the proposed baseline, MECCANO Multimodal leaves space to more investigation on the proposed EHOI detection task due to the challenging nature of the considered industrial-like domain.

\subsection{Action Anticipation}
\label{sec:Action_anticipation}
The goal of the action anticipation task is to predict egocentric actions from an observation of the past (see Figure~\ref{fig:concept_action_ant}). 
Let \begin{math}A = [t_{s}, t_{e}]\end{math} be a video segment, where $t_{s}$ and $t_{e}$ are the start and the end times of the action respectively. The aim is to assign the correct action class $c_i \in C_a$ to the segment $A$ observing a $t_{o}$ (observation time) seconds long video segment preceding the start time of the action $t_{s}$ by $t_{a}$ seconds (anticipation time).
Following \cite{furnari2019rulstm} we used Top-k accuracy and Mean Top-5 Recall as evaluation measures and considered different anticipation times.

\begin{figure*}[]
	\centering
	\includegraphics[width=\textwidth]{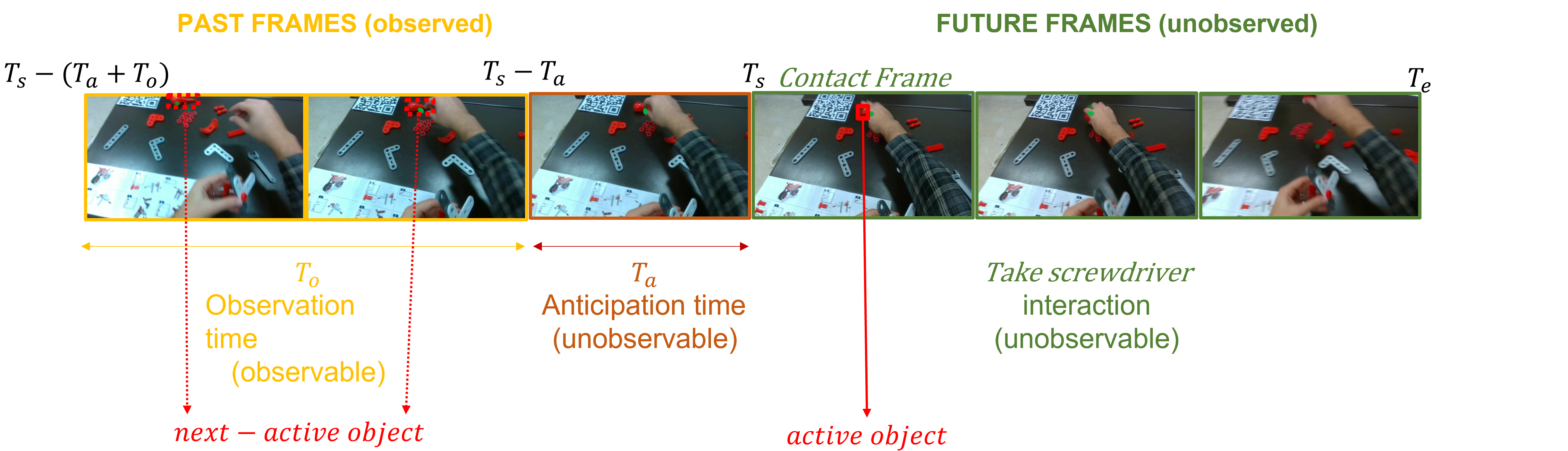}
	\caption{The aim of the Next-Active Object Detection task is to detect and recognize all the objects which will be involved in a future interaction.}
	\label{fig:nao_task}
\end{figure*}

\subsubsection{Method}
We adopted the RULSTM approach proposed in \cite{furnari2019rulstm, furnari2020rulstm} to address the action anticipation task. This model is composed of different branches which take as input different signals (RGB, optical flow and object-centric features). We choose this model due to its state-of-the-art performance and because it has been explicitly designed to work with multimodal observations. We extended this method to exploit the multimodal signals present in the MECCANO dataset. In particular, the adopted baseline is composed of 5 branches, one for each signal: RGB, Depth, Gaze, object-centric features and hands-centric features. 
Depth features have been extracted running SlowFast (Depth) (see Section~\ref{sec:baseline_act_rec} trained on MECCANO. We computed object-centric features following \cite{furnari2019rulstm}, while, gaze features have been obtained weighting the object-centric features with the distance between the center of objects bounding boxes and the gaze position in the image. For hands-centric branch we use the hand annotations of the MECCANO dataset as input.
Branches are trained and fused using the procedures explained in \cite{furnari2019rulstm}. Since we extracted the RGB features using a SlowFast network \cite{feichtenhofer2018slowfast} which encodes both spatial and temporal features, we did not consider the optical-flow. 

\begin{table}[]
\caption{The table reports the ablation study performed with the adopted baseline for action anticipation. All the combinations of 5 branches representing different input signals (RGB, Depth, Objects, Gaze and Hands) are considered.}
\label{tab:ablation_action_ant}
\centering
\resizebox{0.8\columnwidth}{!}{%
\begin{tabular}{c|c|c|c|c|c}
\textbf{RGB} & \textbf{Depth} & \textbf{OBJ} & \textbf{Gaze} & \textbf{Hands} & \textbf{mt5r} \\ \hline
\textbf{\checkmark} & X & X & X & X & 22,88\% \\
X & \textbf{\checkmark} & X & X & X & 14,07\% \\
\textbf{\checkmark} & \textbf{\checkmark} & X & X & X & 14,12\% \\
X & X & \textbf{\checkmark} & X & X & 29,41\% \\
\textbf{\checkmark} & X & \textbf{\checkmark} & X & X & 29,03\% \\
X & \textbf{\checkmark} & \textbf{\checkmark} & X & X & 25,26\% \\
\textbf{\checkmark} & \textbf{\checkmark} & \textbf{\checkmark} & X & X & 21,74\% \\
X & X & X & \textbf{\checkmark} & X & 29,79\% \\
\textbf{\checkmark} & X & X & \textbf{\checkmark} & X & 29,63\% \\
X & \textbf{\checkmark} & X & \textbf{\checkmark} & X & 24,04\% \\
X & X & \textbf{\checkmark} & \textbf{\checkmark} & X & 31,46\% \\
\textbf{\checkmark} & \textbf{\checkmark} & X & \textbf{\checkmark} & X & 29,42\% \\
\textbf{\checkmark} & X & \textbf{\checkmark} & \textbf{\checkmark} & X & 32,01\% \\
\textbf{\checkmark} & \textbf{\checkmark} & \textbf{\checkmark} & \textbf{\checkmark} & X & 28,05\% \\
X & X & X & X & \textbf{\checkmark} & 30,06\% \\
\textbf{\checkmark} & X & X & X & \textbf{\checkmark} & 29,86\% \\
X & \textbf{\checkmark} & X & X & \textbf{\checkmark} & 24,43\% \\
X & X & \textbf{\checkmark} & X & \textbf{\checkmark} & 31,13\% \\
X & X & X & \textbf{\checkmark} & \textbf{\checkmark} & 31,33\% \\
\textbf{\checkmark} & \textbf{\checkmark} & X & X & \textbf{\checkmark} & 29,89\% \\
\textbf{\checkmark} & X & \textbf{\checkmark} & X & \textbf{\checkmark} & 31,31\% \\
\textbf{\checkmark} & X & X & \textbf{\checkmark} & \textbf{\checkmark} & 30,97\% \\
X & \textbf{\checkmark} & \textbf{\checkmark} & X & \textbf{\checkmark} & 27,69\% \\
X & \textbf{\checkmark} & X & \textbf{\checkmark} & \textbf{\checkmark} & 26,84\% \\
X & X & \textbf{\checkmark} & \textbf{\checkmark} & \textbf{\checkmark} & \textbf{32,25\%} \\
\textbf{\checkmark} & \textbf{\checkmark} & \textbf{\checkmark} & X & \textbf{\checkmark} & 29,21\% \\
\textbf{\checkmark} & \textbf{\checkmark} & X & \textbf{\checkmark} & \textbf{\checkmark} & 28,11\% \\
\textbf{\checkmark} & X & \textbf{\checkmark} & \textbf{\checkmark} & \textbf{\checkmark} & 31,30\% \\
X & \textbf{\checkmark} & \textbf{\checkmark} & \textbf{\checkmark} & \textbf{\checkmark} & 27,77\% \\
\textbf{\checkmark} & \textbf{\checkmark} & \textbf{\checkmark} & \textbf{\checkmark} & \textbf{\checkmark} & 24,05\%
\end{tabular}
}

\end{table}

\begin{table}[]
\caption{Results obtained for the action anticipation task considering different values of $t_a$ (anticipation time).}
\label{tab:act_ant}
\resizebox{\columnwidth}{!}{%
\begin{tabular}{lcccccccc}
\multicolumn{1}{c}{\textbf{$t_a$}} & \textbf{2} & \textbf{1.75} & \textbf{1.50} & \textbf{1.25} & \textbf{1} & \textbf{0.75} & \textbf{0.50} & \textbf{0.25} \\ \hline
\multicolumn{1}{l|}{\textbf{Top-1 Acc.}} & 23.37 & 23.48 & 23.30 & 23.97 & 24.08 & 24.50 & 25.60 & 28.87 \\
\multicolumn{1}{l|}{\textbf{Top-5 Acc.}} & 54.65 & 55.99 & 56.56 & 57.73 & 58.23 & 59.96 & 61.31 & 63.40 \\
\multicolumn{1}{l|}{\textbf{M. Top-5 Rec.}} & 18.57 & 18.73 & 21.24 & 21.26 & 22.38 & 24.67 & 24.93 & 26.01
\end{tabular}
}

\end{table}

\subsubsection{Results}
We performed an ablation study considering several instances of the adopted baselines focusing on different combination of these five branches (see Table~\ref{tab:ablation_action_ant}). 
Results show how hard it is to choose a combination of different signals to solve this task. For example, considering the combination of RGB and Depth signals (third row) decreases the performance with respect to using only the RGB signal (first row). Also using all signals simultaneously (last row) does not guarantee best performance, highlighting that it is not sufficient to use all the available signals to solve this task in this challenging environment. We found that the best approach which obtained a Mean Top-5 Recall of 32.25, computed on the Validation Set of the MECCANO dataset, is composed of three branches which take as input the gaze signal, the object-centric features and the hand-centric features. Table~\ref{tab:act_ant} reports the results obtained on the Test set of the MECCANO dataset. We evaluated this baseline considering different anticipation times ($t_a$) ranging from 2~seconds to 0.25 seconds. Qualitative results are shown in Figure~\ref{fig:act_ant_qual}.

\subsection{Next-Active Object Detection}
\label{sec:NAO}
The aim of the Next-Active Object Detection task is to detect and recognize all the objects which will be involved in a future interaction (see Figure~\ref{fig:nao_task}).
Let \begin{math}T = [t_{s}, t_{e}]\end{math} be a EHOI segment, where $t_{s}$ and $t_{e}$ are the start and the end times of the interaction respectively and let \begin{math} O_{act} = \{o_1, o_2, ..., o_n\} \end{math} be the set of active objects. The goal is to predict the set of active objects involved in the interaction $I$, \begin{math} O_I = \{o_1, o_2, ..., o_m\} \end{math} with $O_I \subseteq O_{act}$ and their bounding boxes \begin{math} B_I = \{b_{o_1}, b_{o_2}, ..., b_{o_m}\} \end{math} where $b_{o_i} = (x,y,w,h)$ is the bounding box related to the object $o_i$, by observing a $t_{o}$ (observation time) seconds long temporal segment preceding the start time of the interaction $t_{s}$ by $t_{a}$ seconds (anticipation time).
For evaluation purposes we used the mean Average Precision (mAP) measure which considers the class and the accuracy of the spatial detection of the objects.

\subsubsection{Method}
We explored the task adopting different simple baselines based on the Faster R-CNN object detector \cite{ren2015faster}. The first baseline has been trained using only the active object annotations (the same ones used for the active object detection and recognition task). The second baseline has been trained using only the next-active object annotations. The third baseline has been trained with the active object annotations and finetuned with the next-active objects annotations. The fourth baseline has been trained using both active and next-active objects annotations.

\begin{table}[]
\caption{Results obtained for the Next-active object detection task.}
\label{tab:NAO}
\resizebox{\columnwidth}{!}{%
\begin{tabular}{lcc}
\textbf{Method} & \textbf{mAP} & \textbf{mAP50} \\ \hline
\multicolumn{1}{l|}{Faster R-CNN (active objects)} & \underline{14.10} & \underline{26.00} \\
\multicolumn{1}{l|}{Faster R-CNN (next-active objects)} & 9.90 & 18.20 \\
\multicolumn{1}{l|}{Faster R-CNN (active + finetuning next-active objects)} & 11.60 & 19.90 \\
\multicolumn{1}{l|}{Faster R-CNN (active + next-active objects)} & \textbf{14.20} & \textbf{26.40}
\end{tabular}
}

\end{table}

\begin{figure}[t]
	\centering
	\includegraphics[width=\columnwidth]{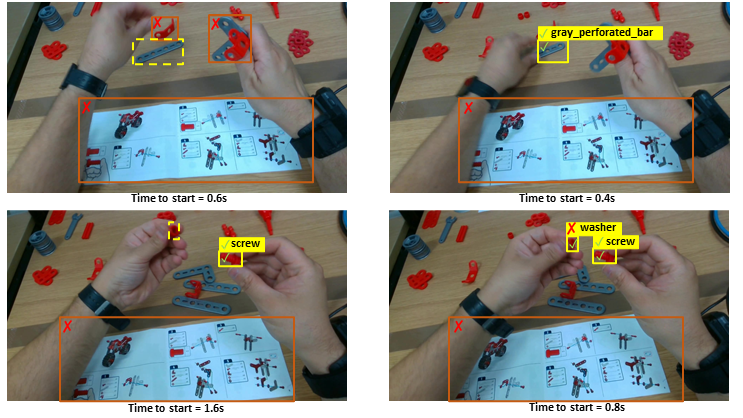}
	\caption{Qualitative results of the best approach based on Faster-RCNN. The dashed bounding box indicates a ground truth object which has not been detected.}
	\label{fig:qual_nao}
\end{figure}

\subsubsection{Results}
Table~\ref{tab:NAO} shows the obtained results. Using only the next-active objects annotations is not enough to obtain reasonable performance on the prediction of the next-active objects. Training the model using both active and next-active objects annotations allows to obtain the best performance considering the mAP (14.20) and the mAP50 (26.40). We reported qualitative results in Figure~\ref{fig:qual_nao}. In general, the task needs to be explored in depth due to the challenging nature of the MECCANO dataset.

\section{Conclusion}
\label{sec:conclusion}
We proposed MECCANO, a multimodal dataset to study egocentric human behavior understanding in an industrial-like scenario. We publicly release the dataset (\url{https://iplab.dmi.unict.it/MECCANO/}) with temporal (action and interaction segments) and spatial (active, next-active object, and hands bounding boxes) annotations considering a taxonomy of 12 verbs, 20 nouns and 61 unique actions. In addition, we performed baseline experiments on five challenging tasks, showing the usefulness of multimodality of the MECCANO dataset. We argue that these multimodal signals are useful to develop real applications to support humans in the real life. 
MECCANO is also suitable to explore different tasks \cite{EgoProceLECCV2022, mixup_active_obj_detction} other than those considered in this work. 
Future works will explore new approaches to improve performance on the proposed tasks.

\section*{\uppercase{Acknowledgements}}
This research is supported by Next Vision\footnote{Next Vision: https://www.nextvisionlab.it/} s.r.l., by MISE - PON I\&C 2014-2020 - Progetto ENIGMA  - Prog n. F/190050/02/X44 – CUP: B61B19000520008, and MIUR AIM - Attrazione e Mobilita Internazionale Linea 1 - AIM1893589 - CUP: E64118002540007

\section*{\uppercase{Supplementary Material}}
\label{sec:supp_material}
This document is intended for the convenience of the reader and reports additional information about the action-interaction relations, the proposed dataset and the annotation stage. This supplementary material is related to the following submission:\\

F. Ragusa, A. Furnari, G. M. Farinella. MECCANO: A Multimodal Egocentric Dataset for Humans Behavior Understanding in the Industrial Domain. submitted to Computer Vision and Image Understanding (CVIU), 2022.  \\
The reader is referred to the manuscript and to our web page \url{https://iplab.dmi.unict.it/MECCANO/} to download the dataset and for further information.

\begin{figure*}[t]
	\centering
	\includegraphics[width=\textwidth]{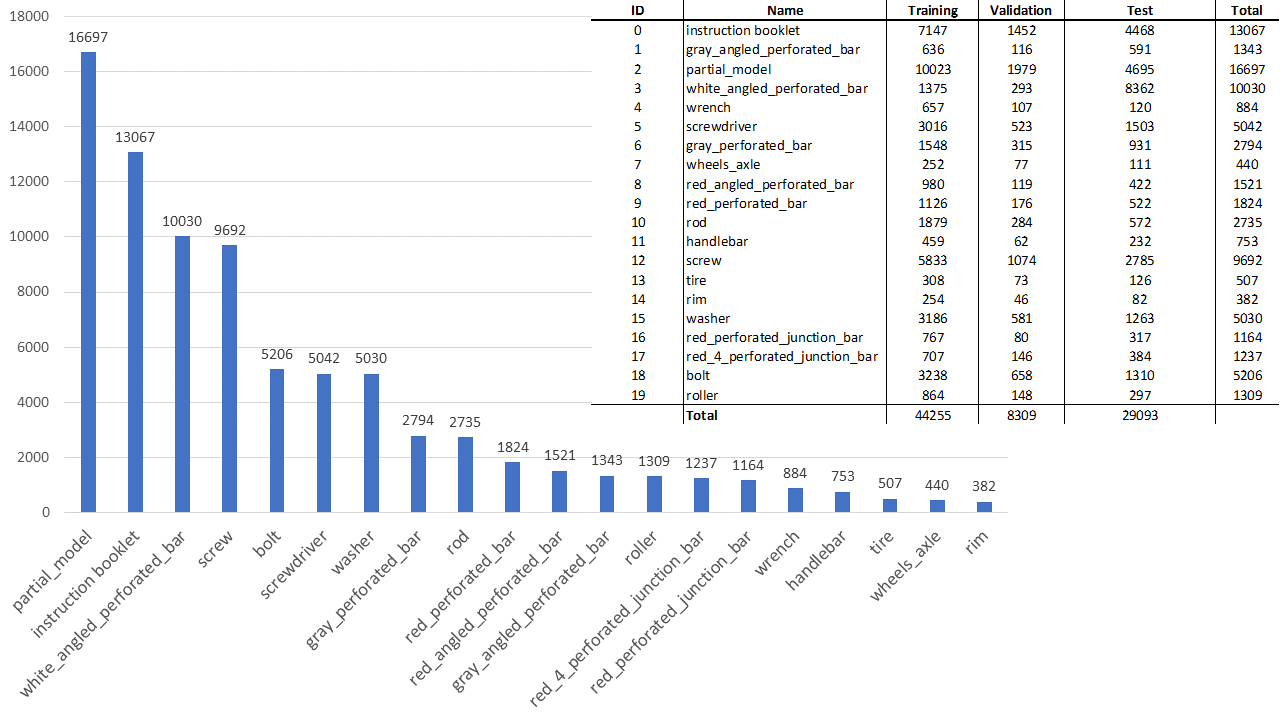}
	\caption{Long-tail distribution of bounding boxes over all object classes.}
	\label{fig:nao_bbox}
\end{figure*}

\begin{figure*}[t]
\centering
\includegraphics[width=\textwidth]{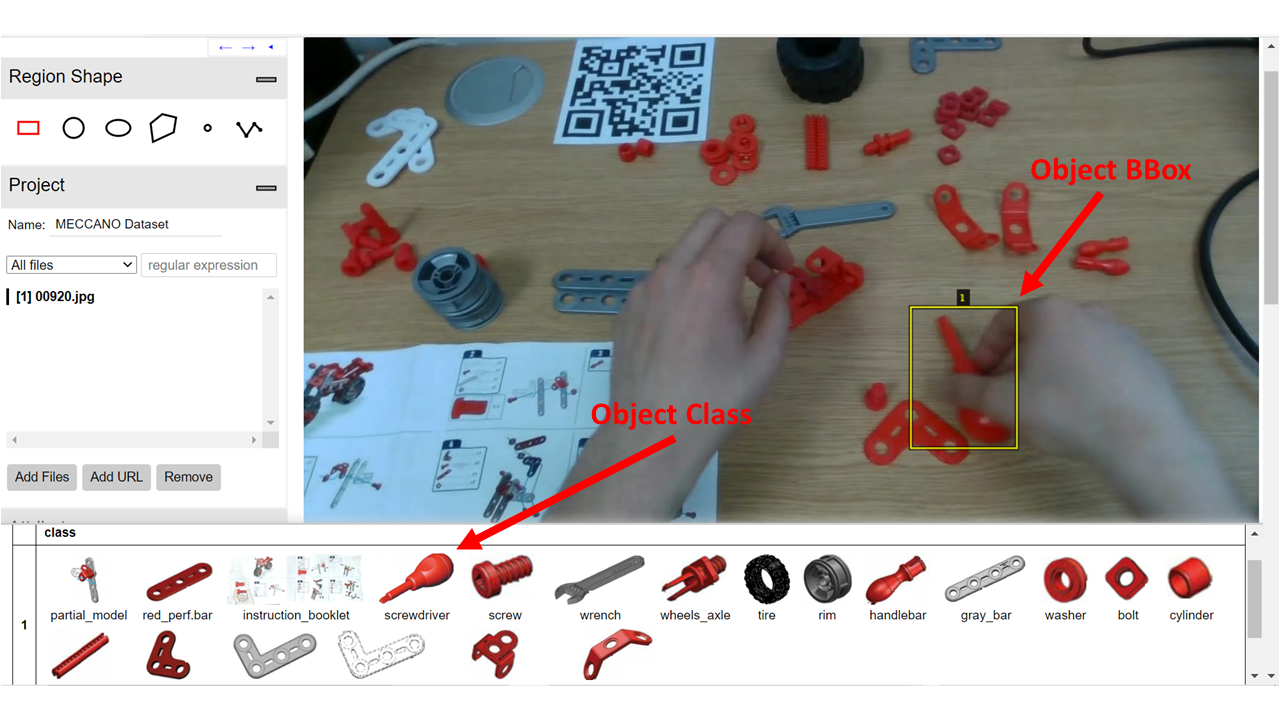}
\caption{Customized VIA project to support the labeling of next-active objects. Annotators were presented with a panel which allowed them to identify object classes through their thumbnails.}
\label{fig:VIA}
\end{figure*}

\section{Next-Active Objects Annotations}
\label{sec:nao}
\begin{figure}[t]
	\centering
	\includegraphics[width=\columnwidth]{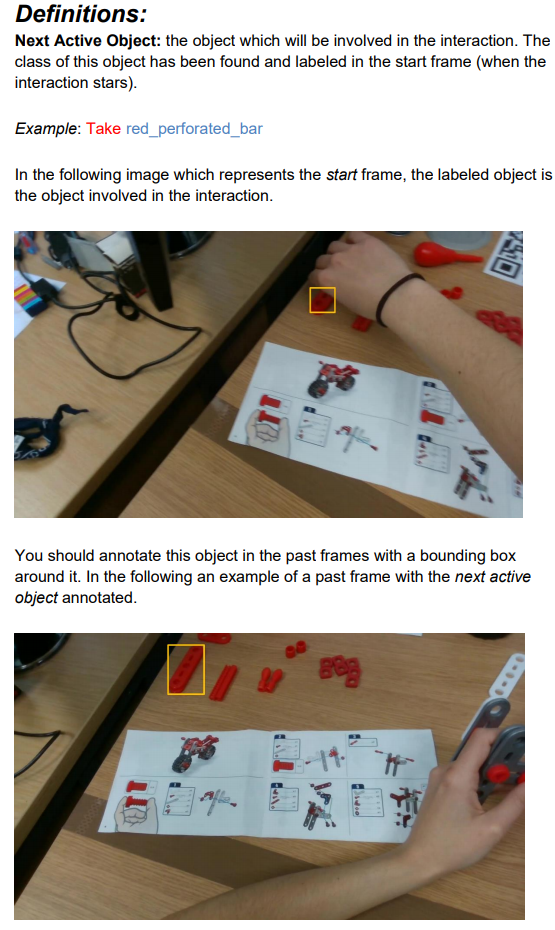}
	\caption{Next active object definition given to the labelers for the next active object bounding box annotation stage.}
	\label{fig:rules_ann}
\end{figure}

Figure~\ref{fig:nao_bbox} shows the distribution of bounding boxes over all object classes. As shown in the figure, the distribution follows a long-tail distribution, which highlights the complexity of this industrial scenario. Moreover, we reported how many bounding boxes we annotated for each object class considering the three splits (Training, Validation and Test) of the dataset.
For the annotation phase of the next-active objects, we used VGG Image Annotator (VIA) \cite{dutta2019vgg} with a customized project to facilitate and speed up the selection of the correct object class (see Figure~\ref{fig:VIA}. Moreover, we provided a document to the annotators, containing a set of key rules for the annotations of next active objects, to support them and reduce ambiguities. In annotation guidelines, we reported the fundamental definitions (e.g., next active object, next active object in a pile, occluded next active object) showing visual examples (see Figure~\ref{fig:rules_ann}). 

\begin{figure}[htp]
	\centering
	\includegraphics[width=\columnwidth]{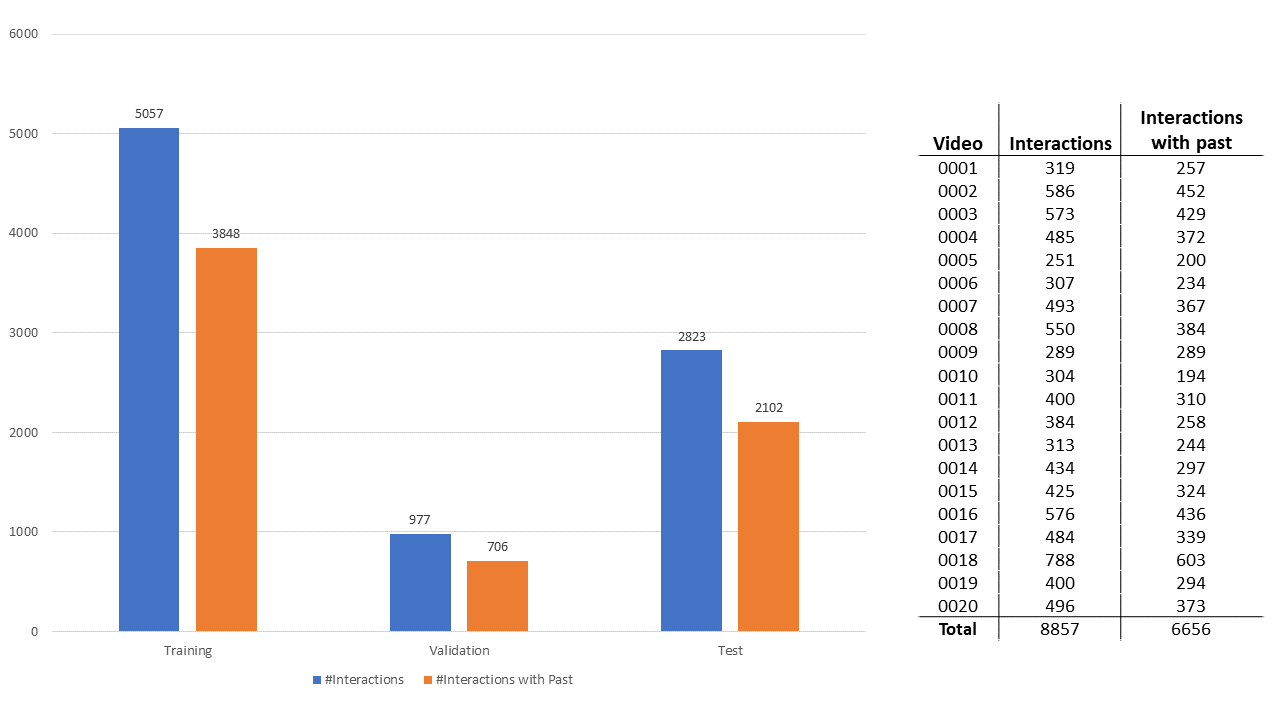}
	\caption{Comparison between the number of interactions with respect to the number of interactions which have past frames.}
	\label{fig:past_interactions}
\end{figure}

Figure~\ref{fig:past_interactions} shows the comparison between the number of interactions present in the MECCANO dataset with respect to the number of interactions which include labeled past frames.

\section{Hands Annotations}
\begin{figure*}[t]
	\centering
	\includegraphics[width=\textwidth]{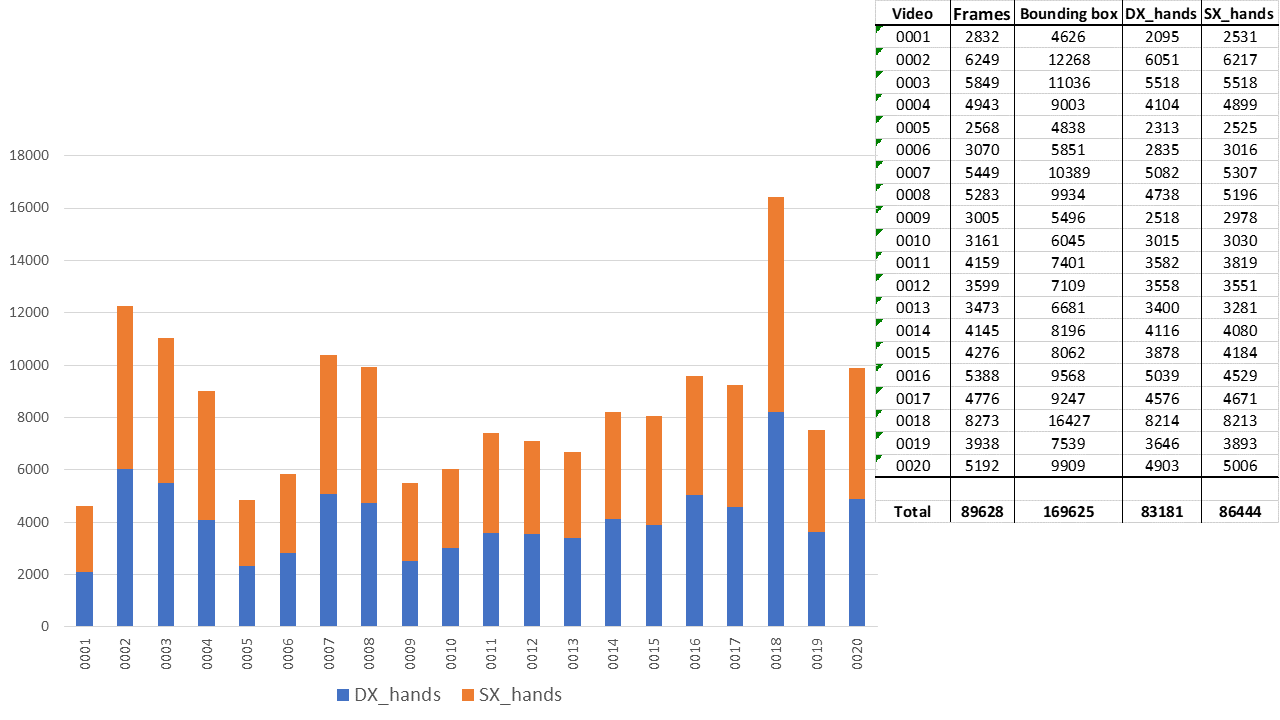}
	\caption{Hands annotations distribution.}
	\label{fig:hands_stats}
\end{figure*}

Figure~\ref{fig:hands_stats} reports some statistics related to the hand annotations.

\begin{figure}[t]
	\centering
	\includegraphics[width=\columnwidth]{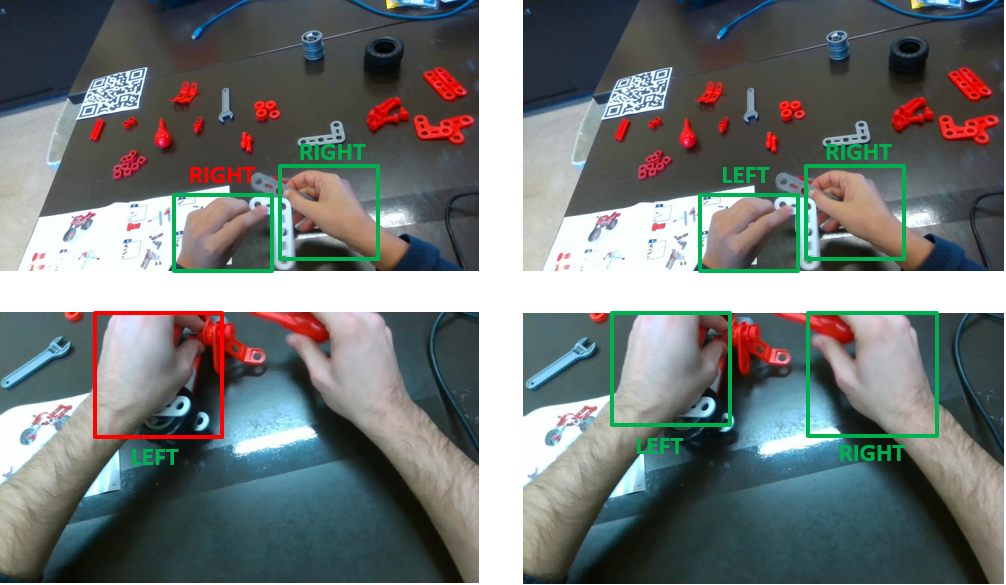}
	\caption{Example of the labeling procedure of the hands.}
	\label{fig:hands_refine}
\end{figure}

An example of the labeling procedure is shown in Figure~\ref{fig:hands_refine}. In the first column, we reported the predictions of the Hand Object Detector \cite{Hands_in_contact_Shan20}. In the second column, the annotators fixed the class errors and refined the bounding box around the hands.

\clearpage
{\small
\bibliographystyle{ieee_fullname}
\balance
\bibliography{ref}
}

\end{document}